# Title page

1. **Type of the article:** Original Research Paper.

2. **Title of the article:** Lost without translation – Can Transformer (language models) understand mood states?
3. **Running title:** Mood representation in Indic language models

4. **Contributors:**

| | Name | Academic qualifications | Department | Institution | City and country |
|---|---|---|---|---|---|
| 1 | Prakrithi Shivaprakash | MD, PDF, DM | Department of Psychiatry | National Institute of Mental Health and Neuro Sciences (NIMHANS) | Bengaluru, India |
| 2 | Diptadhi Mukherjee | MD, PDF, DM | Department of Addiction Medicine | Lokopriya Gopinath Bordoloi Regional Institute of Mental Health (LGBRIMH), Tezpur | Assam, India |
| 3 | Lekhansh Shukla* | MD, PDF, DM | Centre for Addiction Medicine, Department of Psychiatry | National Institute of Mental Health and Neuro Sciences (NIMHANS) | Bengaluru, India |
| 4 | Animesh Mukherjee | B-Tech, M-Tech, PhD (Computer Science and Engineering) | Department of Computer Science and Engineering | Indian Institute of Technology, Kharagpur (IIT-KGP) | Kharagpur, India |
| 5 | Prabhat Chand | MD, DNB, MNAMS | Centre for Addiction Medicine, Department of Psychiatry | National Institute of Mental Health and Neuro Sciences (NIMHANS) | Bengaluru, India |
| 6 | Pratima Murthy | DPM, MD, FRCP (Glasgow) | Centre for Addiction Medicine, Department of Psychiatry | National Institute of Mental Health and Neuro Sciences (NIMHANS) | Bengaluru, India |

5. **Department and Institution to which work must be attributed:** Centre for Addiction Medicine, Department of Psychiatry, NIMHANS, Bengaluru - 560029, India.

6. **Corresponding author***
   - Name: Lekhansh Shukla
   - Full postal address: Office of the Centre for Addiction Medicine, 2nd floor, Centre for Addiction Medicine Ward – Female wing, Department of Psychiatry, National Institute of Mental Health and Neuro Sciences (NIMHANS), Bengaluru – 560029, India.
   - Email address: drlekhansh@gmail.com
   - Telephone number: +91 9886160956.

7. **Counts**
   - Total number of pages: 8 (excluding figures)
   - Word counts
     - for abstract: 249
     - for the text: 2395 (excluding tables, figures, references and acknowledgements)
   - No. of figures: 3
   - No. of tables: 2

8. **Source(s) of support:** This work is supported by an award from Neuromatch, Inc. as part of the Generative AI for Mental Health Research Accelerator, funded by Wellcome Trust Limited. The funder had no role in this study's design, data collection, data analysis, and reporting.

9. **Conflicts of Interest:** The authors have no conflicts of interests to declare.

10. **Acknowledgements:** Nil.

11. **Ethics statement:** The study was approved by the Institute Ethics Committee [NIMHANS/EC (BEH.SC.DIV.)/2025 dated September 3, 2025].

12. **Declaration regarding the use of generative AI:** The authors attest that there was no use of generative artificial intelligence (AI) technology in the generation of text, figures, or tables of this manuscript.

# Lost without translation – Can Transformer (language models) understand mood states?

Prakrithi Shivaprakash, Diptadhi Mukherjee, Lekhansh Shukla, Animesh Mukherjee, Prabhat Chand, Pratima Murthy

## Abstract

**Background:** Large Language Models show promise in psychiatry but are English-centric. Their ability to understand mood states in other languages is unclear, as different languages have their own idioms of distress.

**Aim:** To quantify the ability of language models to faithfully represent phrases (idioms of distress) of four distinct mood states (depression, euthymia, euphoric mania, dysphoric mania) expressed in Indian languages.

**Methods:** We collected 247 unique phrases for the four mood states across 11 Indic languages. We tested seven experimental conditions, comparing k-means clustering performance on: (a) direct embeddings of native and Romanised scripts (using multilingual and Indic-specific models) and (b) embeddings of phrases translated to English and Chinese. Performance was measured using a composite score based on Adjusted Rand Index, Normalised Mutual Information, Homogeneity and Completeness.

**Results:** Direct embedding of Indic languages failed to cluster mood states (Composite Score = 0.002). All translation-based approaches showed significant improvement. High performance was achieved using Gemini-translated English (Composite=0.60) and human-translated English (Composite=0.61) embedded with gemini-001. Surprisingly, human-translated English, further translated into Chinese and embedded with a Chinese model, performed best (Composite = 0.67). Specialised Indic models (IndicBERT and Sarvam-M) performed poorly.

**Conclusion:** Current models cannot meaningfully represent mood states directly from Indic languages, posing a fundamental barrier to their psychiatric application for diagnostic or therapeutic purposes in India. While high-quality translation bridges this gap, reliance on proprietary models or complex translation pipelines is unsustainable. Models must first be built to understand diverse local languages to be effective in global mental health.

## Key words

Deep learning, Mood disorders, Machine translation, Multilingualism, India, Idioms of distress.

# Introduction

Applications of artificial intelligence (AI) in psychiatric research and practice have attracted substantial attention. They are being studied as diagnostic as well as therapeutic tools [1]. This entire technological revolution followed the development of transformer models, introduced in the 2017 paper "Attention Is All You Need" [2]. A transformer model is a type of neural network architecture that is designed to handle sequential data like sentences or paragraphs, by understanding the context and relationships between different parts of the sequence. This was achievable by a mechanism called "self-attention", which allowed models to weigh the importance of different words in a sentence relative to each other, leading to a deeper understanding of language as used by humans. A cornerstone of these models is the ability to represent words along with their context as semantic embeddings. For example, consider this sentence: "Ajay didn't go to office today because he felt too tired." The model builds a numerical representation of each word, along with its meaning and context. It understands that "Ajay" is the word corresponding to "he" in this particular context. It simultaneously creates a web of relationships between "Ajay", "office", and "tired", capturing the full meaning of the sentence and giving relevant attention to each word (for a detailed explanation of this, read [3]). A Large Language Model (LLM) is built using this transformer architecture and is trained on massive amounts of data, which helps it understand, predict, and generate human-like language. Examples of state-of-the-art models are OpenAI's Generative Pretrained Transformer-5 (GPT-5) [4] and Google DeepMind's Gemini-2.5-Pro [5].

A familiar analogy is that, embedding models (also known as encoders), for example, Bidirectional Encoder Representations from Transformers (BERT) [6] are equivalent to Wernicke's area (processes incoming words and extracts underlying meaning – language comprehension). The subsequent LLM revolution is akin to connecting these comprehension centres with the Broca's area (language generation – articulating the words, generating correct sentences) and the prefrontal cortex (planning and reasoning to synthesise the right information). For example, it is claimed that transformer models encode clinical knowledge – a version of Pathways Language Model (PaLM), called Med-PaLM generated answers to medical questions, with performance comparable to human clinicians [7].

However, the quality of embeddings and the performance of LLMs depend on the language. Since these models are trained predominantly on English, their ability to represent (encode) and manipulate other languages is not the same as English [8]. Psychiatry depends heavily on language – it is the most valuable component of eliciting psychiatric symptomatology.  An important task in psychiatric applications is the ability to distinguish between mood states. This is typically done during a standard psychiatric interview by asking the patient how their mood has been over the past two weeks. In addition to observations on affect, speech and thought, the response to this question is used to diagnose pathological mood states like depression, dysphoric or euphoric mania.

In the machine learning (ML) community, this task is referred to as emotion recognition, and transformer models like Bidirectional Encoder Representations from Transformers (BERT) perform exceptionally well for detecting emotions in text [6].

In this research, we investigate the capacity of current transformer models to represent textual responses associated with mood states accurately. There are two key factors to consider here. First, patients frequently use idioms of distress instead of direct references to sadness or happiness [9]. Second, translation of such emotion-laden text is not a trivial matter. Research has shown that machine translation flattens and, at times, changes the emotion [10].

# Methods

We aim to quantify the performance of current state-of-the-art transformer models in faithfully representing mood states when these states are expressed in Indian languages. The data for this study were collected from Mental Health Professionals (MHPs), persons with lived experience of mental illness and caregivers through a dedicated survey portal on the project's website (https://heads-ai.com/datacollection). This was advertised via social media and a targeted campaign. The targeted campaign involved sharing the survey link within professional networks of psychiatrists and psychiatry residents, as well as with lived experience experts known to the research team. We did not collect any identifying information or proof of being an MHP, and there was no restriction on the number of responses per respondent. Quality control is, however, maintained after data collection, where we only took unique and relevant responses.

We asked the respondent to type in Roman script or native script. There were four questions, each corresponding to four mood states: euthymia, depression, dysphoric mania or euphoric mania. The exact question depended on the responder's selected role; for example, for an MHP, it would be "How do clinically depressed patients respond to the following question in your native language – how has your mood been for the past two weeks?". Figure 1 shows the study design. The study received ethical clearance from the Institute Ethics Committee.

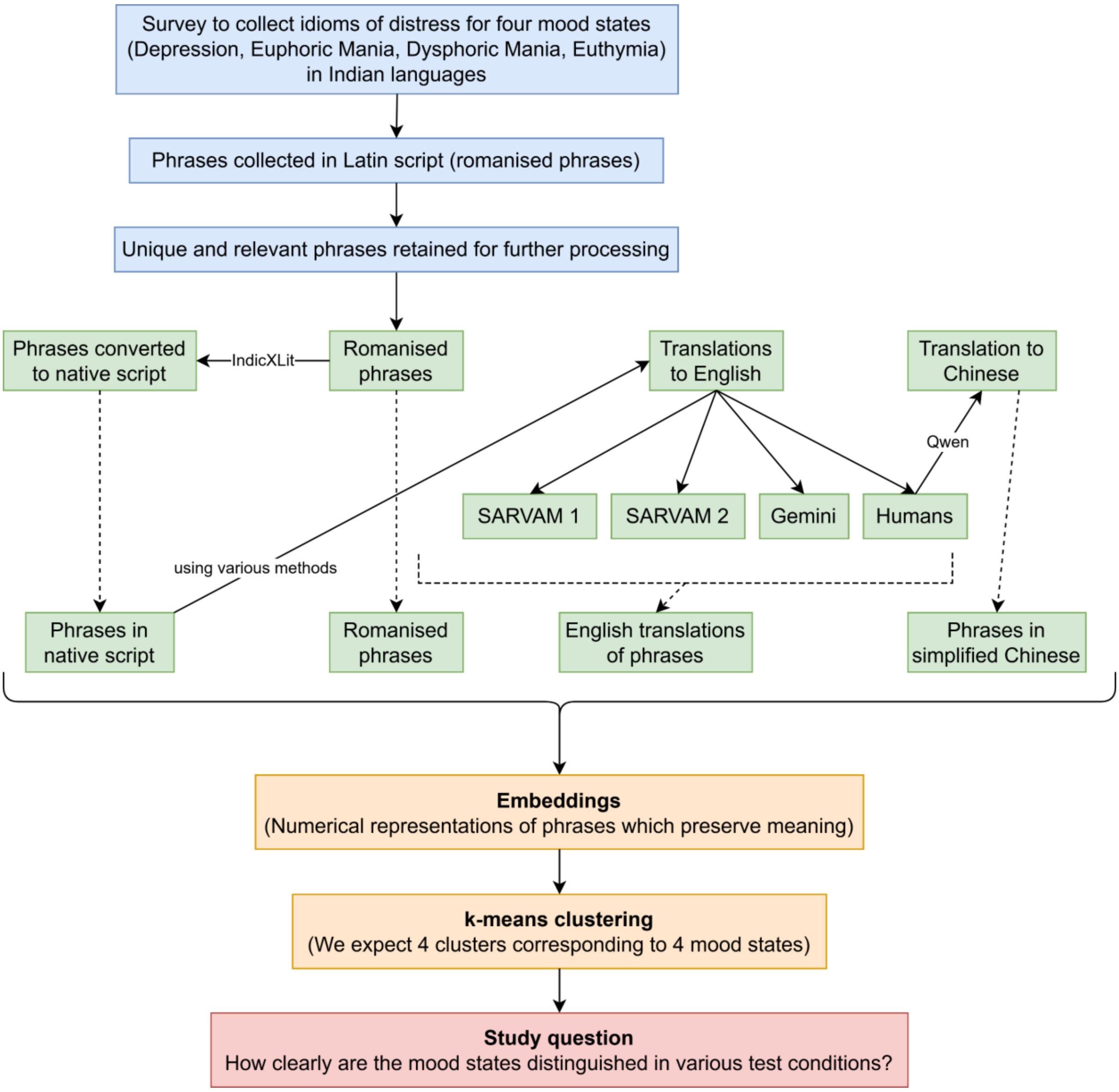

*Figure 1: Study design and methodology*

A priori, these phrases belong to one of the four predefined clusters – euthymia, depression, euphoric mania or dysphoric mania. We hypothesise that if a model faithfully represents these phrases, a k-means clustering of the embeddings must also give four distinct clusters. We examine the following experimental settings (Table 1). We note that the models listed in the Input column (IndicXLit, AI4Bharat Indic to English models, Sarvam-M, Gemini-2.5-Pro, Qwen-3-Max) are used for translation of phrases, and those listed in the Embedding Models column are used to create the semantic embeddings for each phrase.

*Table 1: Experimental conditions with various input and embedding model combinations.*

| Condition | Input | Embedding Models |
|---|---|---|
| 1 | Phrases in Indian Languages but written in Roman script | paraphrase-multilingual-MiniLM-L12-v2 and gemini-001 |
| 2 | Phrases in Indian Languages in their native scripts, generated from roman script entries with IndicXLit [11]. | indicBERT and gemini-001 |
| 3 | Phrases translated to English using the Sarvam Translate service (which uses AI4Bharat Indic to English models) | paraphrase-multilingual-MiniLM-L12-v2 and gemini-001 |
| 4 | Phrases translated to English using Sarvam-M, an instruction-tuned LLM optimised for Indic languages. | |
| 5 | Phrases translated to English using Gemini-2.5-Pro. | |
| 6 | Phrases translated to English by bilingual psychiatrists | |
| 7 | English versions of the phrases obtained above are translated to Chinese (simplified) using Qwen-3-Max | Qwen-3-embedding-8b |

It is essential to distinguish between conditions 3 and conditions 4 and 5. In conditions 4 and 5, we prompt an LLM to translate with specific instructions to preserve emotional content and relevant context. In contrast, in condition 3, we use machine translation, which cannot be adapted to a particular context.

The clustering performance is evaluated with Adjusted Rand Index (ARI), Normalised Mutual Information (NMI), Homogeneity, and Completeness [12]. ARI measures how closely the model's predicted labels match the true labels for a given phrase, adjusted for chance matches – i.e., agreement between the model's clusters of mood states and the ground truth (scored -1=strong disagreement, 0=random agreement, 1=perfect agreement). NMI gives certainty of the model's clustering being closer to ground truth (0 to 1, with 1 indicating perfect clustering). Homogeneity measures the extent to which each cluster contains members of only one class, while completeness measures the extent to which a class are assigned to the same cluster (for both, 0 to 1 for perfect clustering). For all these measures, we need true labels and predicted labels. The predicted label is the majority label of a cluster. The true labels are the four pre-defined mood states. Finally, we report classification accuracy which is the proportion of phrases that are correctly classified into four mood states, for the best and worst performers.

Given these multiple performance metrics, we construct a composite score to rank the conditions. This composite score is a weighted average (ARI: 50%, NMI: 40%, Homogeneity: 5%, Completeness: 5%).

# Results

We received a total of 736 responses; however, only 254 of these were unique by language, mood state and exact string matching. Out of these 254 responses, we removed seven phrases as they were irrelevant to the task. A majority of these (228) are contributed by responders who identified themselves as MHP. As a result, we have 247 unique phrases split across the four mood states (depression: 75, euphoric mania: 60, euthymia: 58 and dysphoric mania: 54). Eleven languages spoken in India are represented in this dataset (Bangla: 73, Hindi: 66,

Kannada: 33, Assamese 17, Tamil 13, English and Malayalam 11, Telugu 7, Khasi 8, Manipuri and Punjabi 4). The dataset is available in the Supplementary Materials.

We tested seven different experimental conditions, evaluating multiple embedding models on their ability to cluster the 247 phrases into the four predefined mood states. A ranking of tested combinations is given in Table 2.

*Table 2: Clustering performance in ascending order for various inputs and embedding model combinations.*

| **Input; Embedding model** | **Composite score** | **ARI** | **NMI** |
|---|---|---|---|
| Phrases in native Indian scripts; IndicBERT | 0.002 | 0.0004 | 0.004 |
| Phrases in roman script; gemini-001 | 0.003 | 0.002 | 0.005 |
| Phrases translated to English with Sarvam-M and indic LLM; gemini-001 | 0.107 | 0.074 | 0.140 |
| Phrases translated to English with Sarvam end-to-end indic to English translator; gemini-001 | 0.114 | 0.085 | 0.143 |
| Phrases translated to English by Gemini-2.5-Pro; gemini-001 | 0.601 | 0.588 | 0.614 |
| Phrases translated to English by bilingual domain experts; gemini-001 | 0.616 | 0.605 | 0.626 |
| Phrases translated to English by bilingual domain experts and then translated to Chinese (simplified) by Qwen-3-Max; Qwen-3-embedding-8b | 0.671 | 0.661 | 0.682 |
| ARI: Adjusted Rand Index; NMI: Normalised Mutual Information | | | |

Without translation (Conditions 1 and 2), no meaningful clustering was possible, regardless of the embedding model. Classification accuracy using native script and IndicBERT embedding model is 31%. Direct embedding of Romanised or native script yielded composite scores near zero, meaning the models completely failed to separate mood states. This indicates existing multilingual and Indic-specific models cannot correctly encode the semantic content of these idioms of distress and meaningfully represent mood states.

In contrast, all translation-based approaches (Conditions 3-7) significantly improved performance, with performance dependent on both the translation and embedding models. Using gemini-001 embeddings, human translation (Condition 6; accuracy 81%) slightly outperformed Gemini translation (Condition 5; accuracy 80%). Surprisingly, the highest performance came from Condition 7 (accuracy 84%), where human-translated English text was subsequently translated to Chinese and embedded using Qwen-3-embedding-8b. The strong performance of Condition 5 indicates that high-quality machine translation can approach the efficacy of human expert translation for this specific task. Figure 2 shows qualitative examples of mood state clustering by the worst and best performing conditions. Supplementary table S1 provides examples and performance in other test conditions.

| Phrase romanised to English (Native language) | English translation | True label | Using native script and IndicBERT | Using Chinese translation and Qwen |
|---|---|---|---|---|
| tumba bejaar agutte (Kannada) | I've been feeling very upset | 😢 | 🙂 ❌ | 😢 ✅ |
| bahot chidchidaahat ho rahi hai (Hindi) | I've been feeling very irritable | 😠 | 😢 ❌ | 😠 ✅ |
| furti lagchhe (Bengali) | I am feeling exuberant | 😁 | 😢 ❌ | 😁 ✅ |
| nga biang (Khasi) | I'm alright | 🙂 | 😢 ❌ | 🙂 ✅ |
| Pretty stable for a change (English) | Pretty stable for a change | 🙂 | 😢 ❌ | 🙂 ✅ |
| Mējāja garama āchē ēkhana. (Bengali) | I'm in a foul mood right now | 😠 | 😢 ❌ | 😢 ❌ |
| bahut pareshan hai mann (Hindi) | My mind has been very troubled | 😢 | 😢 ✅ | 😠 ❌ |
| manasu chanagide (Kannada) | My mind feels good | 🙂 | 🙂 ✅ | 😁 ❌ |
| 🙂 Euthymic | 😢 Depression | 😁 Euphoric mania | 😠 Dysphoric mania | |

*Figure 2: Examples of phrases and their clustering in the worst and best performing conditions*

There was a striking difference in performance between embedding models. For instance, in Condition 6 (Human Translation), gemini-001 embeddings outperformed sentence transformer embeddings. This pattern held across all conditions tested with both, possibly due to gemini-001's larger embedding dimension (3072 vs. 768), which allowed for a more nuanced representation of mood-state differences in idioms of distress.

Indic transformer models, such as IndicBERT [13], which are claimed to be specialised for handling Indian languages, performed much worse than gemini-001. Sarvam-based translations (Conditions 3 and 4) were better than no translation, but they significantly lagged behind those based on the Gemini, Human, and Qwen approaches. In general, closed-source Gemini embeddings outperformed open-source models. Supplementary Table S2 details the performance of all combinations.

Figure 3 shows the confusion matrix heatmaps of cluster compositions when using native script embeddings and translated script embeddings. Note the mixing of different mood state idioms (columns) in all four clusters (rows) in Figure 1a. In contrast, Figure 1b has a well-separated cluster for each mood state. We see that the separation between conditions is substantially better with translated scripts. Supplementary Figures S1 to S12 detail the cluster composition across conditions.

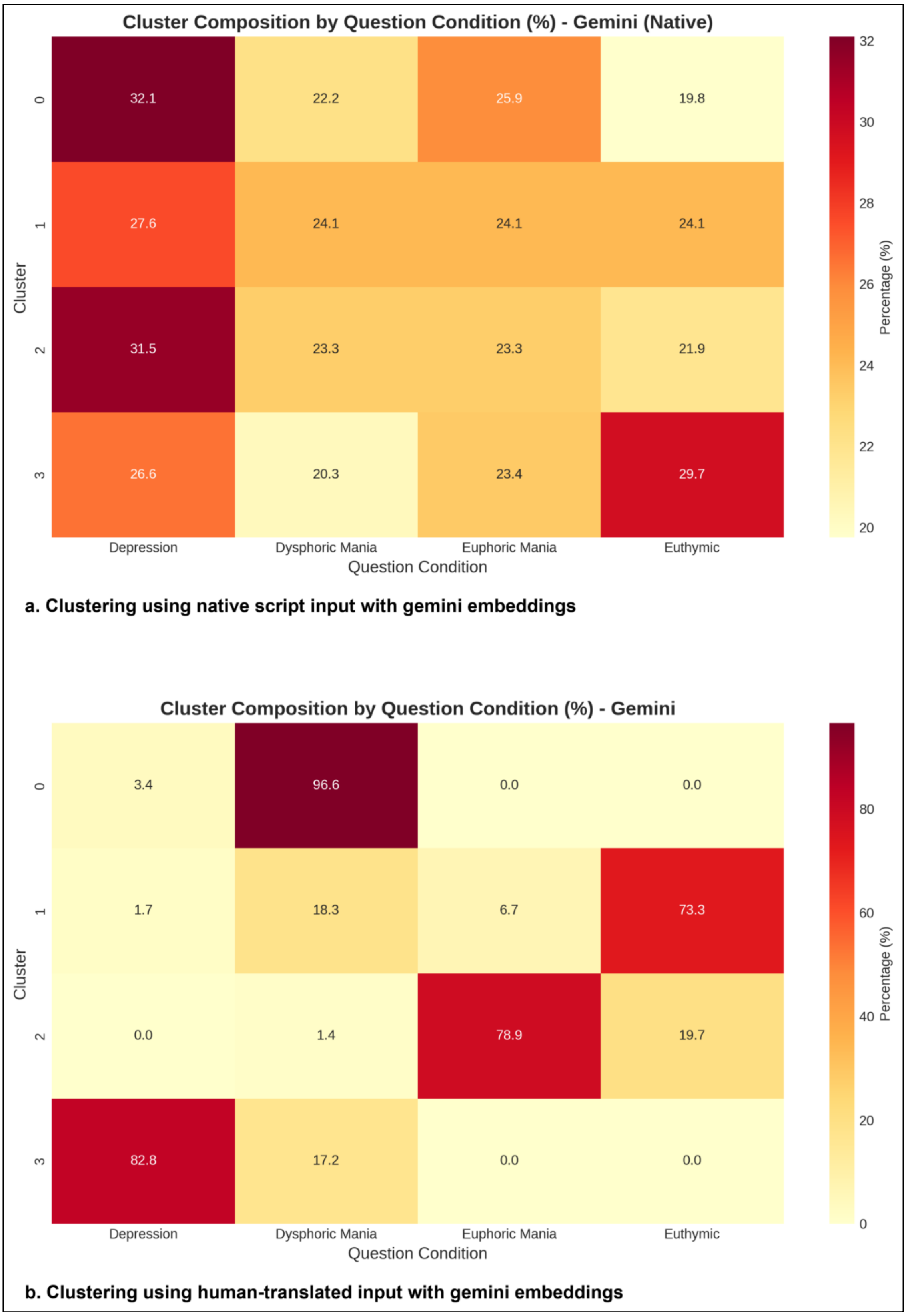

**a. Clustering using native script input with gemini embeddings**

**b. Clustering using human-translated input with gemini embeddings**

*Figure 3: Cluster composition heatmaps with native script and human-translated inputs using Gemini embeddings.*

# Discussion

We found that the existing transformer models cannot faithfully represent pathological mood states expressed directly in Indian languages. However, translation into another language significantly improved representation. Finally, to illustrate the need for developing our own transformer models, which can encode emotions in Indian languages, we chose the Qwen series of models. These open-source models from Chinese research groups excel at translating English to Chinese. Having received substantial training in Chinese, they also excel in encoding emotions in the Chinese language. As a result, we find that the convoluted path of translating Indian language phrases to English and then to Chinese yielded the best performance, likely because it combined high-fidelity human translation (from Indic to English) with a model (Qwen) that has a native, deep understanding of the Chinese language.

While India's Economic Survey (2024-25) highlights the potential of AI and technology for mental healthcare [14], our core finding that current models fail to represent Indic languages presents a fundamental barrier – these AI tools are overwhelmingly English-centric. This has implications for both the mental healthcare community as well as AI research community. The often cited fact that India has one of the world's largest English-speaking populations is confusing, as this figure stems from a large population denominator; the 2011 Census of India indicates that only ~11% of the population speaks English, mainly as a second language [15]. The small, English-speaking fraction (~11%) may successfully use these tools, creating a false impression of "readiness", that masks inaccessibility for the ~90% who communicate in Indic languages. This can risk exacerbating the existing health inequities – it is well known that English proficiency is linked to higher socio-economic status [16].

Specialised models, such as IndicBERT and translation services like Sarvam-M, although optimised for Indic languages, performed poorly. If these models fail to distinguish four distinct and explicit mood states from idioms of distress, it is far-fetched to expect them to understand the implicit cues of a real clinical interaction related to mental health.

The "Transformer" architecture was introduced in 2017 [2], BERT [17] and subsequently, Generative Pre-trained Transformers (GPT) in 2018 [18]. Yet, nearly a decade later, foundational semantic understanding for low-resource languages remains a hurdle. A substantial gap remains to be closed before these models can be considered safe or effective for Indic languages. Many works from the AI research community aim to reach high goals, such as “No language left behind” [19] and “Towards Leaving No Indic Language Behind” [20]. However, for Indic languages, datasets used for training are created from sources such as Wikipedia and online news articles [21]. These formal corpora lack the real-world, vernacular, and emotionally rich data required to represent nuances in idioms of distress, which are central to mental health assessment.

This study has limitations. Our dataset of 247 phrases, although unique, is small and was obtained via crowdsourcing rather than direct clinical interviews. However, for the given task, this limitation does not affect the reliability of our findings. The only relevant point is the use of valid phrases to describe mood states, which has been ensured prior to modelling. We are sharing the dataset for others to examine and critique as well. The distribution of languages was also uneven. However, the dominance of Hindi and Bengali should have boosted the performance of Indic models as these are mid-resource languages. Therefore, this imbalance cannot explain our findings. Despite these limitations, our results highlight two potential paths forward. The first is to develop new, culturally and linguistically grounded models from scratch, trained on diverse, real-world psychiatric clinical data. This is a long-term, resource-intensive goal. The second, more immediate path is to rely on high-quality translation systems. However, our study shows that this, too, has challenges. The open-source Indic translation models failed to provide adequate semantic preservation. We achieved good performance only by using closed-source, paid models, such as Gemini, or the complex Indic-English-Chinese translation path. This reliance on proprietary technology presents its own challenges for building scalable, equitable, explainable and accessible systems for underserved populations.

# Conclusion

Our findings demonstrate that current transformer models, including those specialised for Indic languages, fail to encode mood states expressed in Indic languages semantically. This is a fundamental barrier to their psychiatric application in diverse populations in India. While this semantic gap can be bridged by translating to a high-resource language, relying on proprietary models or complex translation pipelines is financially and computationally unsustainable in low-resource healthcare settings. Before language models can be leveraged for mental health in India, we must first ensure that they can speak and communicate in the local languages of the people.

# Source(s) of support

This work is supported by an award from Neuromatch, Inc. as part of the Generative AI for Mental Health Research Accelerator, funded by Wellcome Trust Limited. The funder had no role in this study's design, data collection, data analysis, and reporting.

# Supplementary file

## Contents

## Confusion matrix heatmaps

### Figure S1: Clustering using romanised text with gemini embeddings.

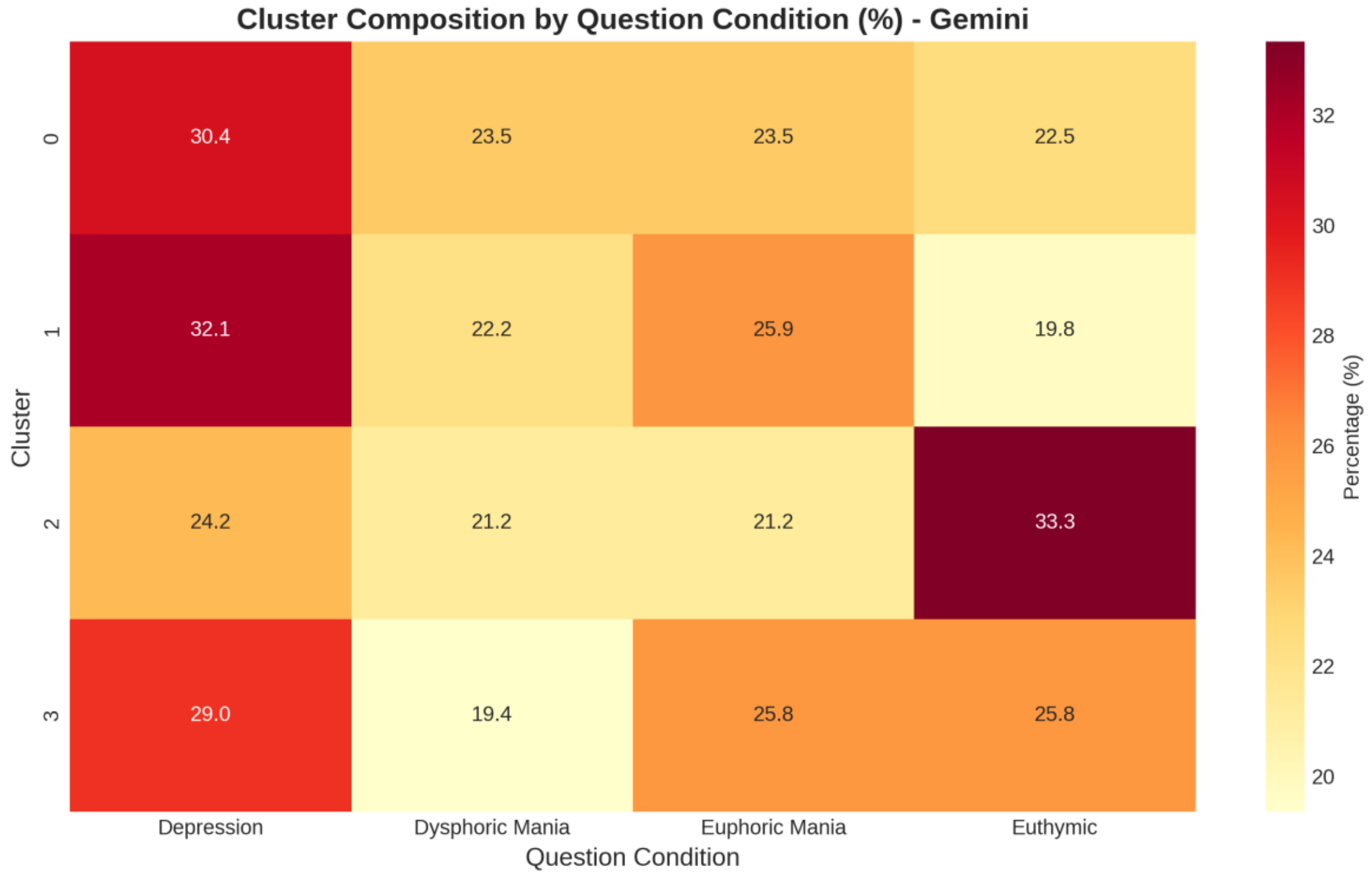

### Figure S2: Clustering using romanised text with sentence_transformer embeddings.

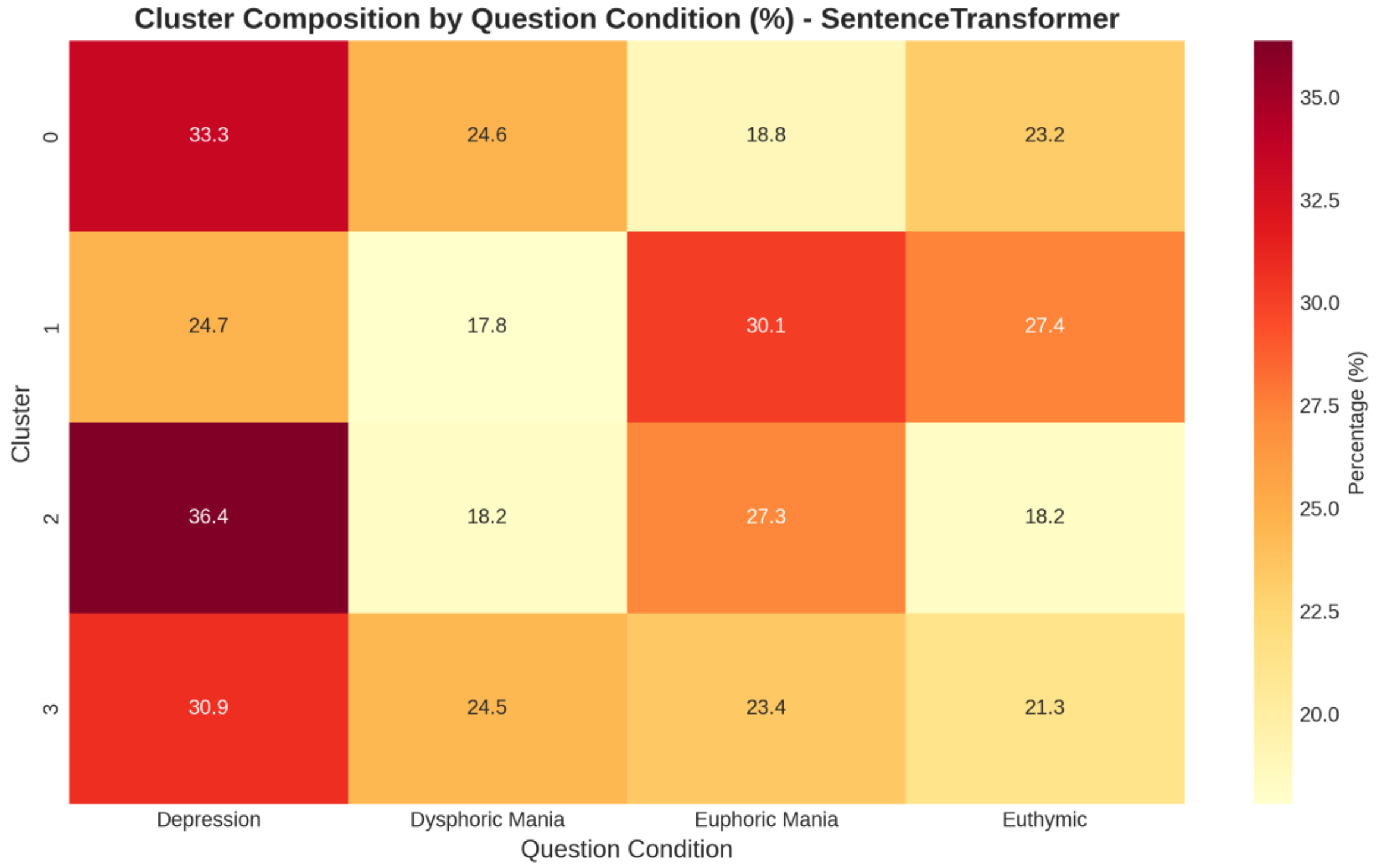

## Figure S3: Clustering using native script with IndicBERT embeddings.

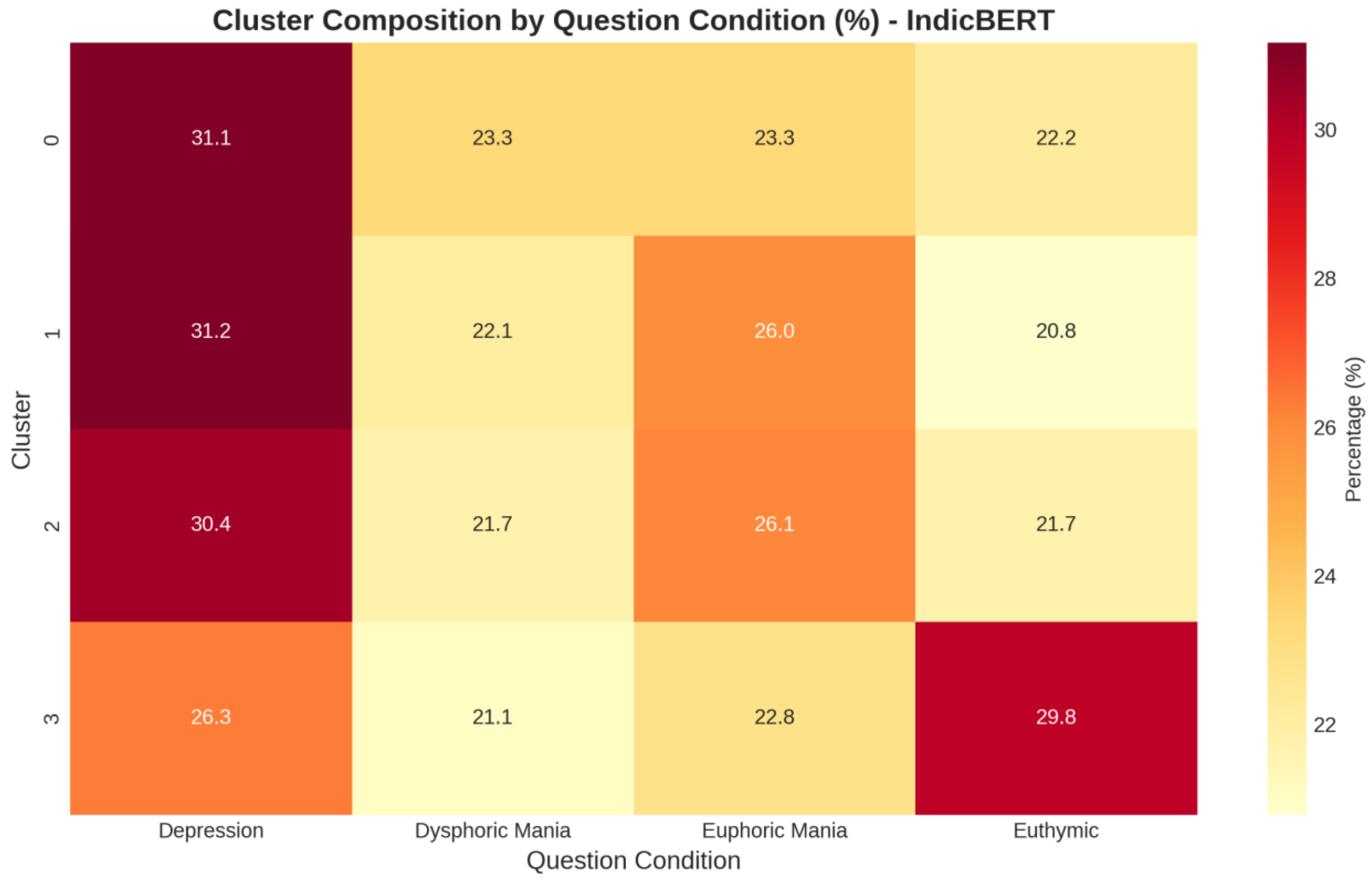

## Figure S4: Clustering using Sarvam indic2en translation with gemini embeddings.

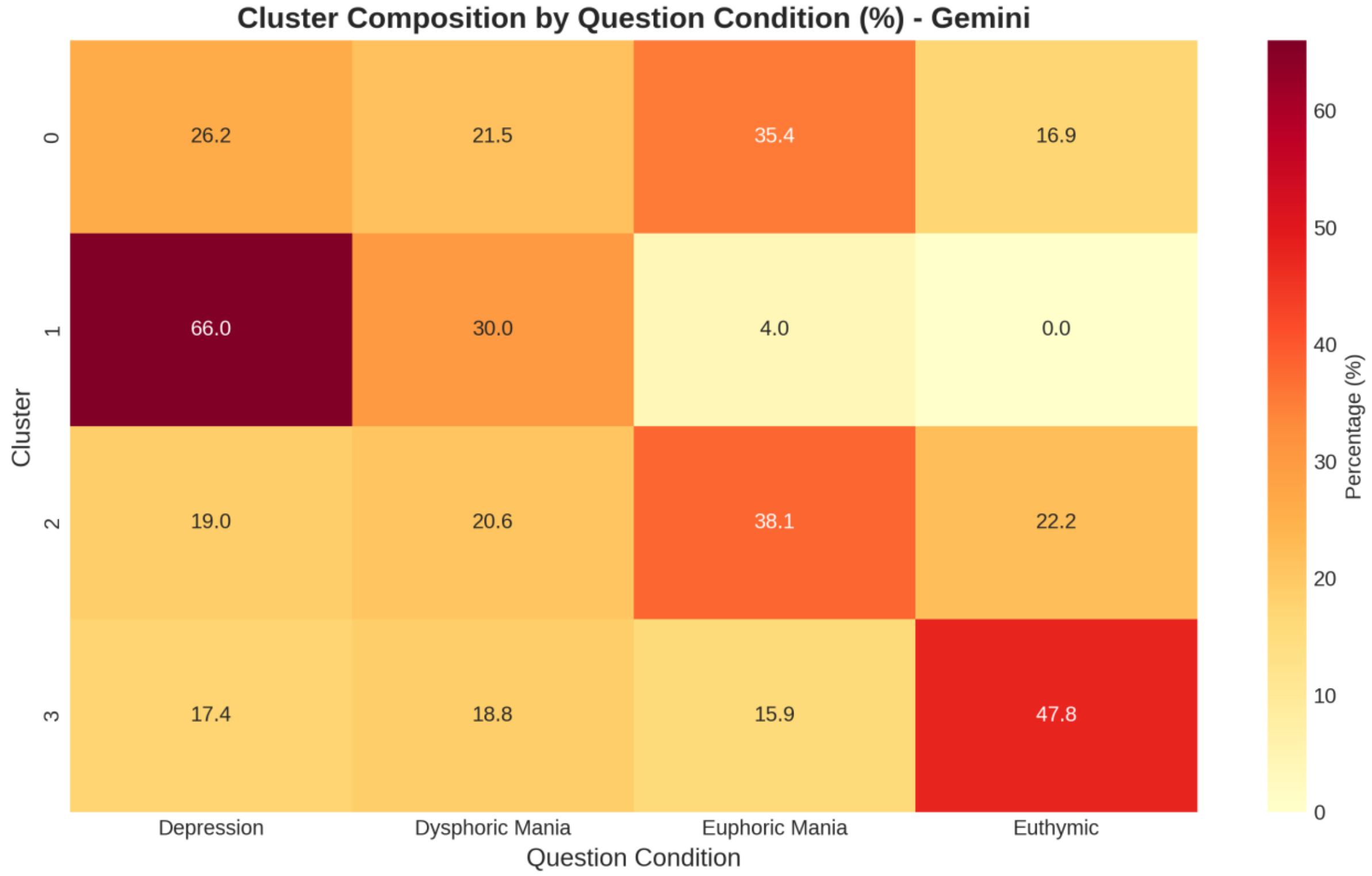

## Figure S5: Clustering using Sarvam indic2en translation with sentence_transformer embeddings.

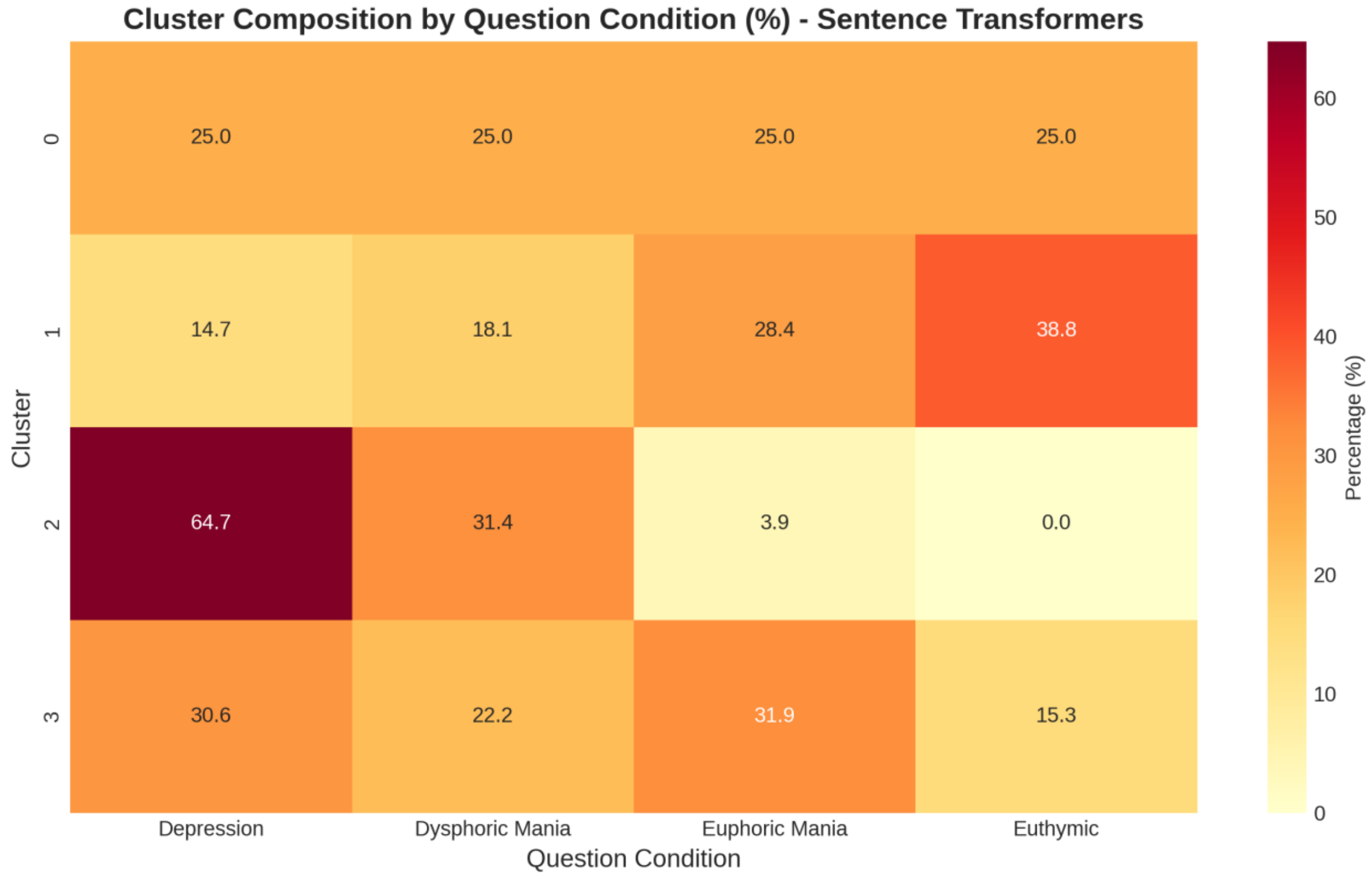

## Figure S6: Clustering using Sarvam-M LLM Translation with gemini embeddings

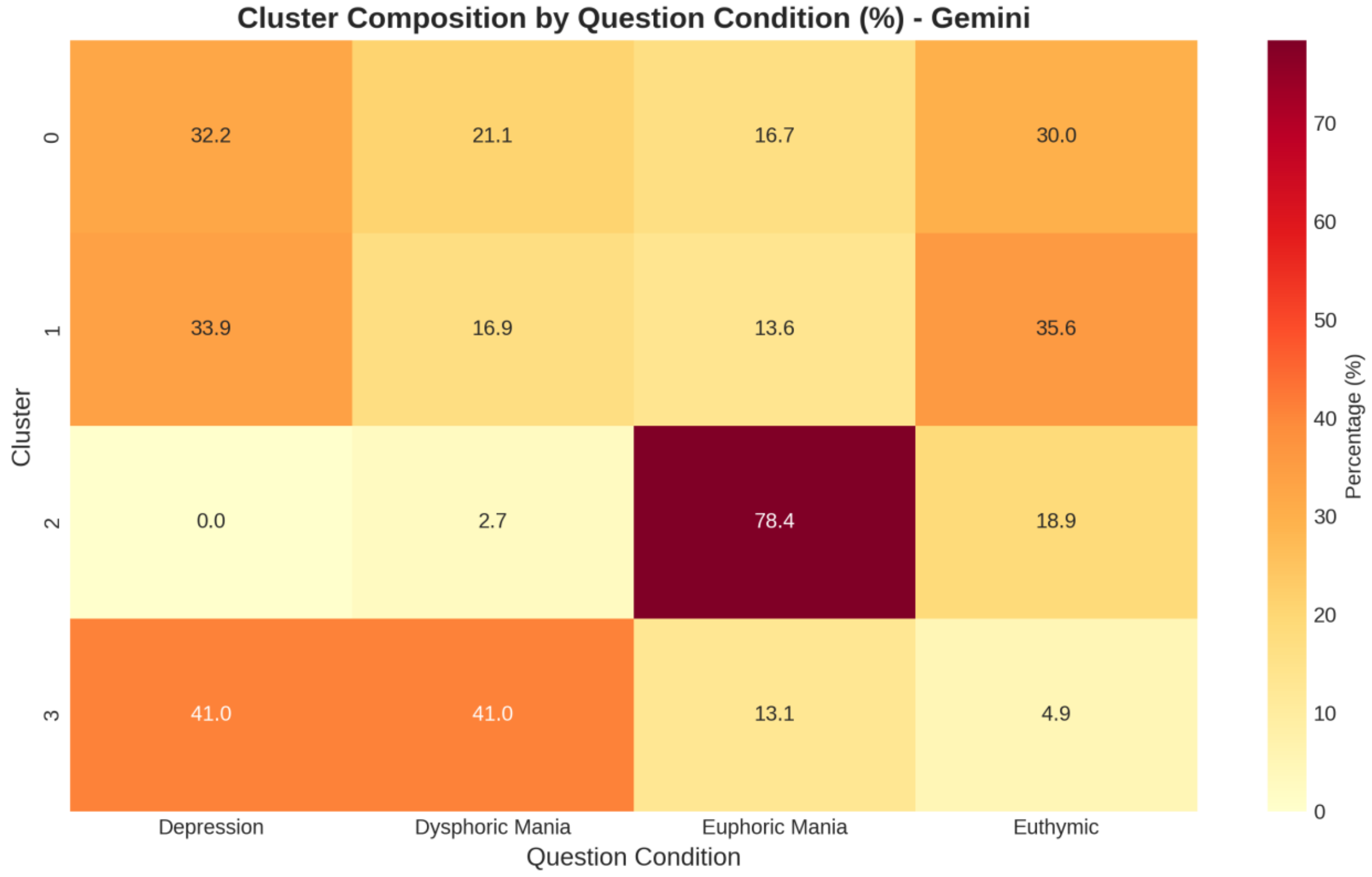

## Figure S7: Clustering using Sarvam-M LLM Translation with sentence_transformer embeddings

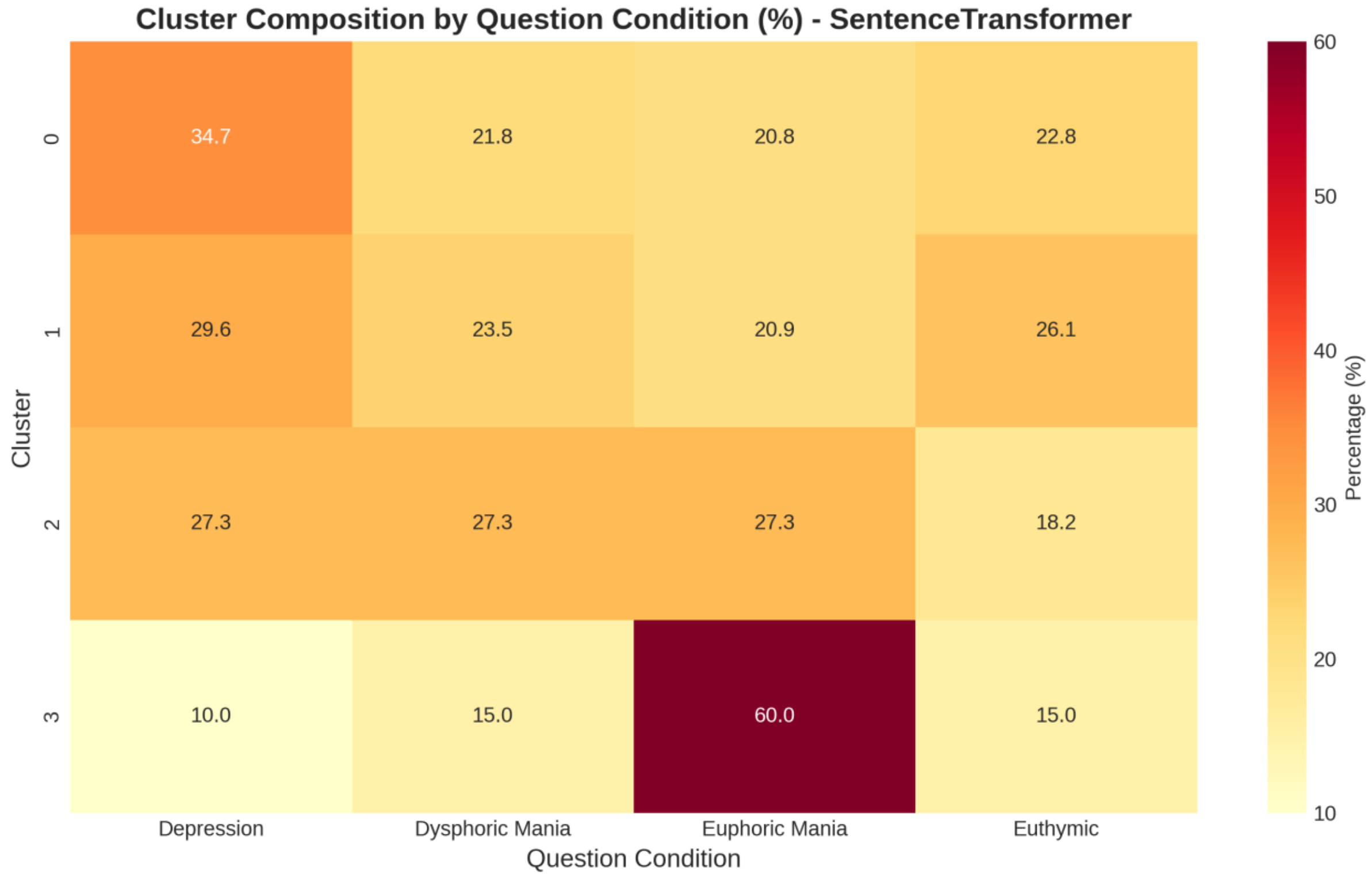

## Figure S8: Clustering using Gemini-2.5-Pro translations with gemini embeddings

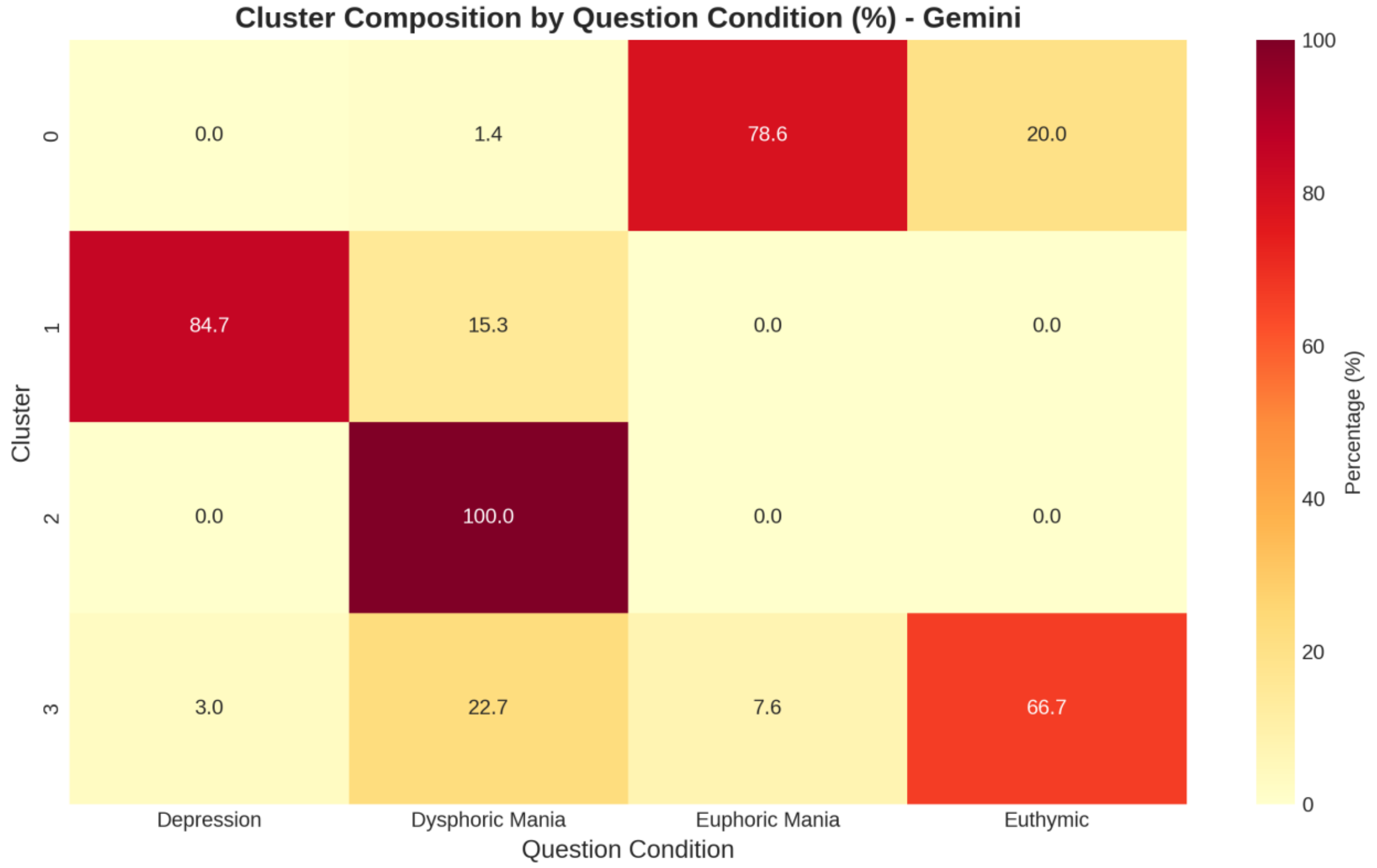

## Figure S9: Clustering using Gemini-2.5-Pro translations with sentence_transformer embeddings

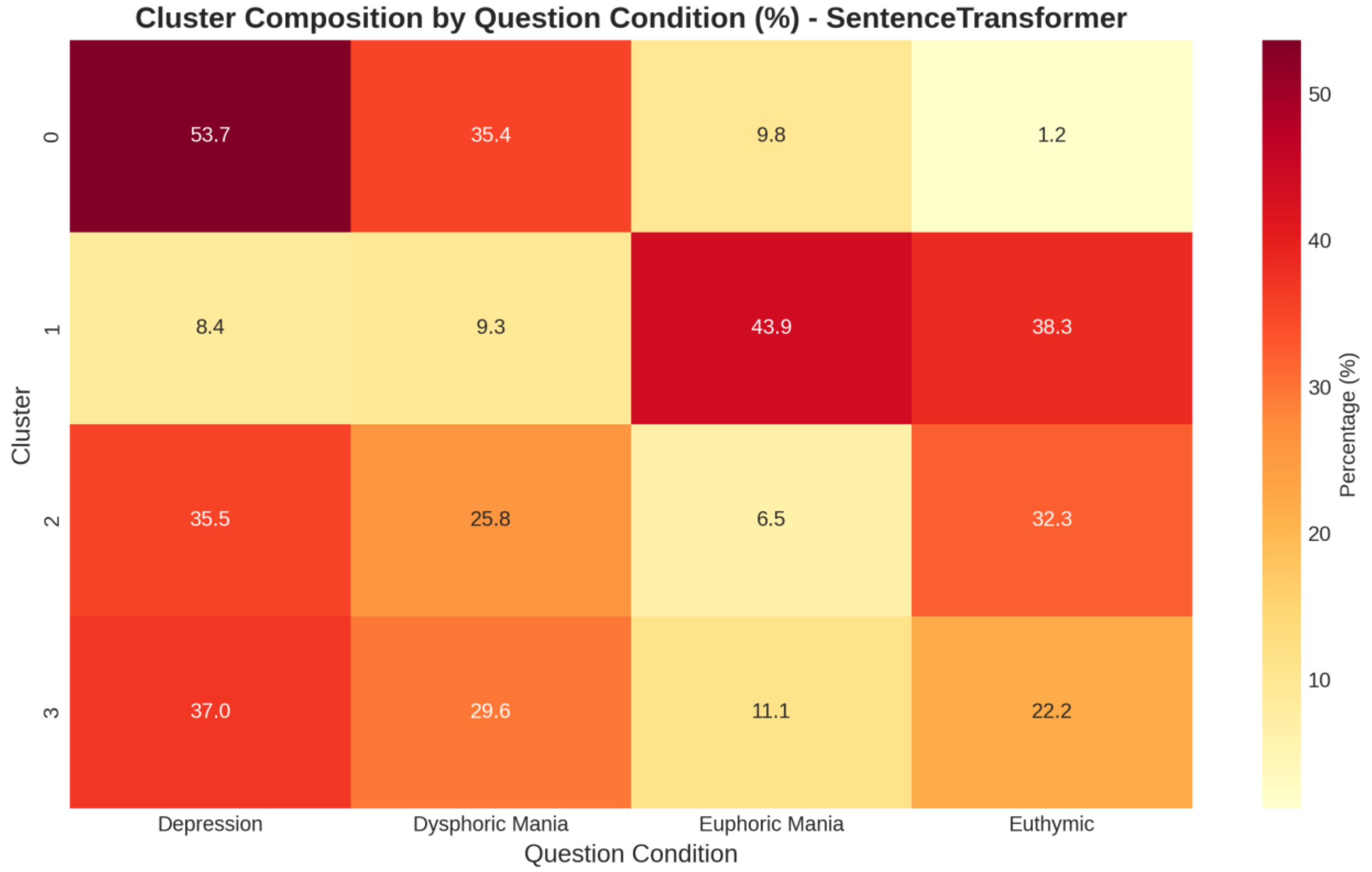

## Figure S10: Clustering using Human English translations with gemini embeddings

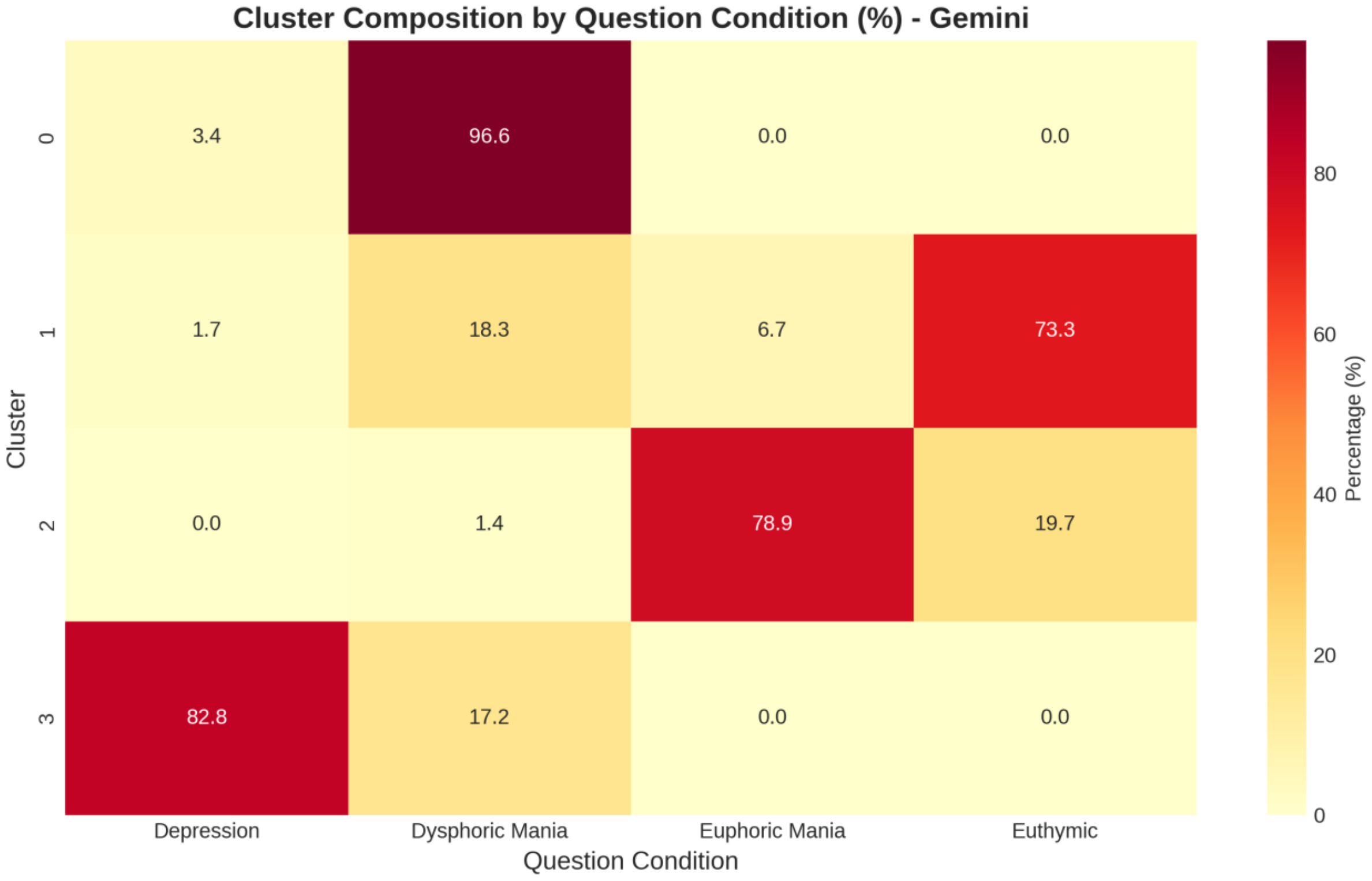

## Figure S11: Clustering using Human English translations with sentence_transformer embeddings

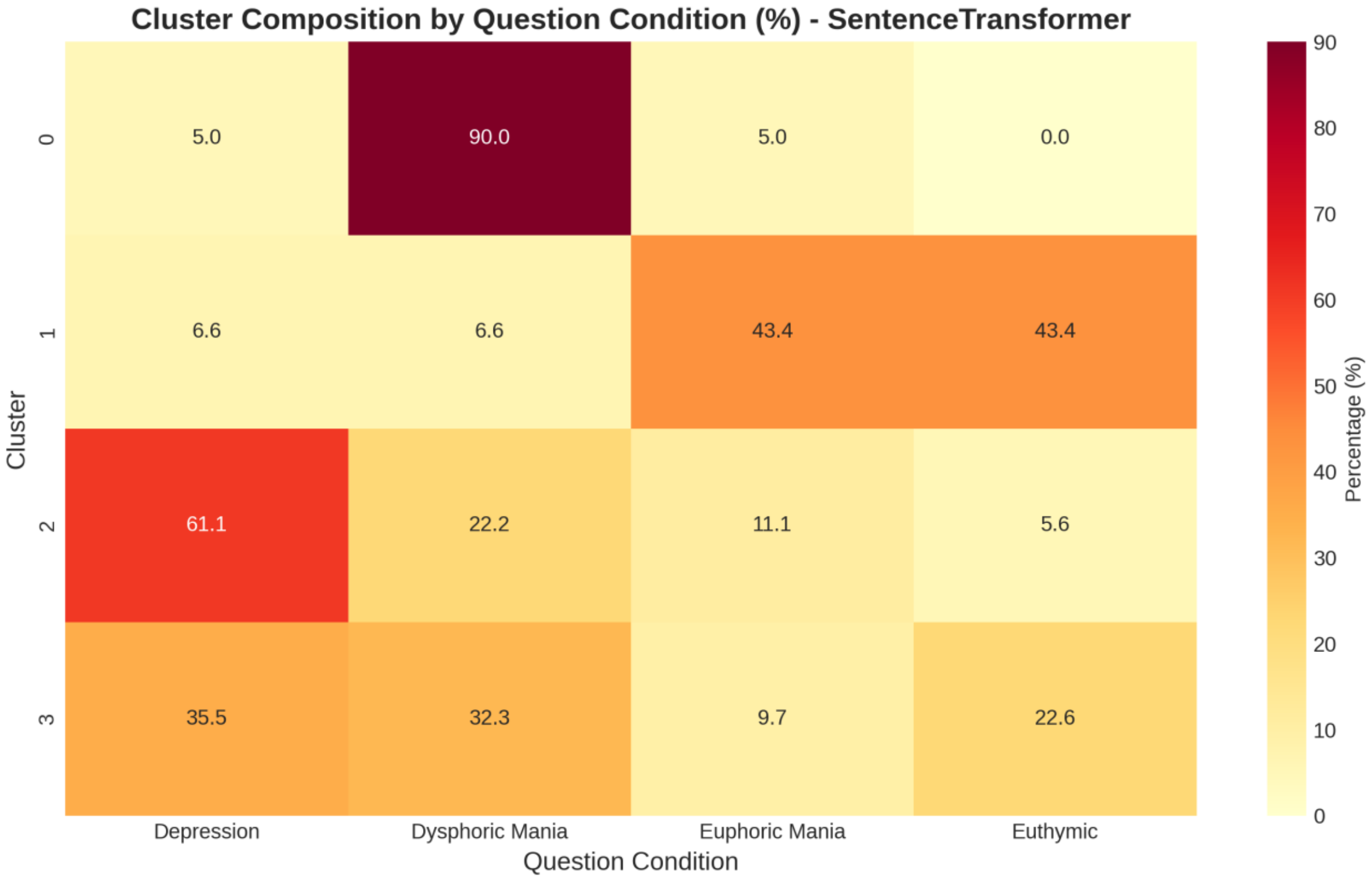

## Figure S12: Clustering using Chinese translation by Qwen with Qwen3-8B embeddings

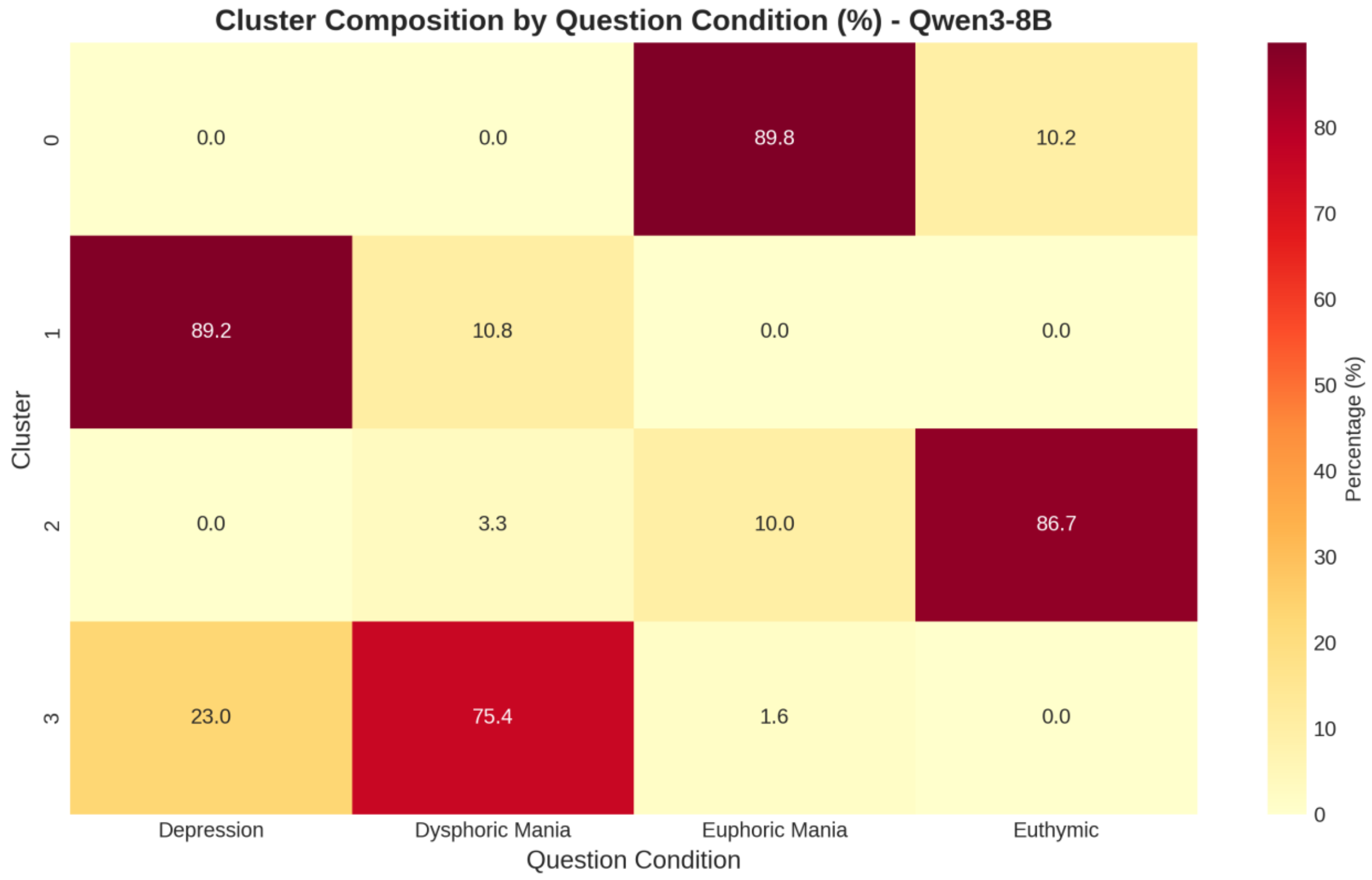

## Table S1: Qualitative examples of phrases and their clustering in various conditions

| Language | Native language | Romanised phrase | Human English translation | True mood state label | Condition 2 Predicted label | Condition 5 Predicted label | Condition 6 Predicted label | Condition 7 Predicted label |
|---|---|---|---|---|---|---|---|---|
| hindi | बोहत अचा है | bohat acha hai | It's been very good. | Euthymic | Depression | Euphoric Mania | Euphoric Mania | Euphoric Mania |
| hindi | मान एकदुम बढ़िया है | mann ekdum badhiya hai | Feeling absolutely great. | Euphoric Mania | Depression | Euphoric Mania | Euphoric Mania | Euphoric Mania |
| hindi | बहोत चिड़चिड़ाहट हो रही है | bahot chidchidaa hat ho rahi hai | I've been feeling very irritable. | Dysphoric Mania | Depression | Dysphoric Mania | Dysphoric Mania | Dysphoric Mania |
| hindi | मान ठीक है | mann theek hai | My spirits are okay. | Euthymic | Depression | Euthymic | Euthymic | Euthymic |
| kannada | ತುಂಬಾ ಬೇಜಾರ್ ಆಗುತ್ತೆ | tumba bejaar agutte | I've been feeling very upset. | Depression | Euthymic | Depression | Depression | Depression |
| bengali | ফুর্তি লাগছে | furti lagchhe | I am feeling exuberant | Euphoric Mania | Depression | Euphoric Mania | Euphoric Mania | Euphoric Mania |
| bengali | ভালো আছি | bhalo aachhi | I've been good. | Euthymic | Depression | Euthymic | Euthymic | Euthymic |
| assamese | বঢ়িয়া | borhiya | Excellent | Euphoric Mania | Depression | Euphoric Mania | Euphoric Mania | Euphoric Mania |
| Manipuri | wakhal se yam khoirang khoisao nei. | wakhal se yam khoirang khoisao nei. | My mind has been very restless and agitated. | Dysphoric Mania | Depression | Dysphoric Mania | Dysphoric Mania | Dysphoric Mania |
| Khasi | nga biang | nga biang | I'm alright. | Euthymic | Depression | Euthymic | Euthymic | Euthymic |
| hindi | चिड़चिड़ा | chidchida | Irritable | Dysphoric Mania | Depression | Dysphoric Mania | Dysphoric Mania | Dysphoric Mania |
| bengali | একদম ভালো, দারুণ ভালো↵↵ফিলিং ভেরি ওয়েল | ekdom bhalo, darun bhalo↵↵feeling very well | Absolutely good, wonderfully good. Feeling very well. | Euphoric Mania | Depression | Euphoric Mania | Euphoric Mania | Euphoric Mania |
| hindi | बोहत खुश है, पहले कभी एआईएसई नही लगा | bohat khush hai, pehle kabhi aise nahi laga | I'm very happy, I've never felt like this before. | Euphoric Mania | Depression | Euphoric Mania | Euphoric Mania | Euphoric Mania |
| Malayalam | അടിപൊളി | adipoli | Fantastic | Euphoric Mania | Euthymic | Euphoric Mania | Euphoric Mania | Euphoric Mania |

| Language | Native language | Romanised phrase | Human English translation | True mood state label | Condition 2 Predicted label | Condition 5 Predicted label | Condition 6 Predicted label | Condition 7 Predicted label |
|---|---|---|---|---|---|---|---|---|
| TAMIL | என் மனசு ரோம்பா வலிக்குது | en manasu rombah valikkuthu | My heart hurts so much. | Depression | Euthymic | Depression | Depression | Depression |
| English | pretty stable for a change | pretty stable for a change | pretty stable for a change | Euthymic | Depression | Euthymic | Euthymic | Euthymic |
| bengali | মুড আমার কদিন ঠিকই আছে | Muḍa āmāra kadina ṭhika'i āchē | My mood has been fine for the past few days. | Euthymic | Depression | Euthymic | Euthymic | Euphoric Mania |
| bengali | মেজাজ গরম আছে এখন। | Mējāja garama āchē ēkhana. | I'm in a foul mood right now. | Dysphoric Mania | Depression | Dysphoric Mania | Dysphoric Mania | Depression |
| TAMIL | என் மனசல இத தங்கிக்கவா முடில | en manasala itha thangikkavah mudila | My mind can't bear this. | Depression | Euthymic | Depression | Depression | Dysphoric Mania |
| Telugu | బాగుంది | baagundi | It's been good. | Euthymic | Depression | Euthymic | Euthymic | Euphoric Mania |
| kannada | ಮನಸು ಚನಾಗಿದೆ | manasu chanagide | My mind feels good. | Euthymic | Euthymic | Euphoric Mania | Euphoric Mania | Euphoric Mania |
| hindi | बहुत परेशन है मान | bahut pareshan hai mann | My mind has been very troubled. | Depression | Depression | Depression | Depression | Dysphoric Mania |
| kannada | ಮನಸ್ಸು ಖುಷಿಯಾಗಿದೆ. | Manas'su khuṣiyāgide. | My heart has been happy. | Euthymic | Euthymic | Euphoric Mania | Euphoric Mania | Euphoric Mania |

## Table S2: Clustering performance ranked for all input and embedding model combinations.

| Rank | Condition | Embedding | ARI | NMI | Homogeneity | Completeness | Composite |
|---|---|---|---|---|---|---|---|
| 1 | C7 | qwen3_8b | 0.6810 | 0.6815 | 0.6827 | 0.6803 | 0.6712 |
| 2 | C6 | gemini | 0.6054 | 0.6263 | 0.6130 | 0.6401 | 0.6159 |
| 3 | C5 | gemini | 0.5883 | 0.6144 | 0.5998 | 0.6297 | 0.6014 |
| 4 | C6 | sentence_transf | 0.2452 | 0.2802 | 0.2331 | 0.3511 | 0.2639 |
| 5 | C5 | sentence_transf | 0.1842 | 0.1926 | 0.1440 | 0.2904 | 0.1909 |
| 6 | C3 | gemini | 0.0846 | 0.1428 | 0.1242 | 0.1679 | 0.1140 |
| 7 | C4 | gemini | 0.0738 | 0.1397 | 0.1168 | 0.1738 | 0.1073 |
| 8 | C3 | sentence_transf | 0.0825 | 0.1126 | 0.0994 | 0.1298 | 0.0977 |
| 9 | C4 | sentence_transf | 0.0094 | 0.0330 | 0.0199 | 0.0975 | 0.0238 |
| 10 | C1 | sentence_transf | 0.0038 | 0.0082 | 0.0059 | 0.0134 | 0.0062 |
| 11 | C1 | gemini | 0.0017 | 0.0046 | 0.0030 | 0.0105 | 0.0034 |
| 12 | C2 | gemini_native | 0.0003 | 0.0039 | 0.0027 | 0.0066 | 0.0022 |
| 13 | C2 | indicbert | 0.0004 | 0.0035 | 0.0024 | 0.0062 | 0.0020 |

| language | question_condition | response_normalized | response_romanized | response_native_script | human_translation |
|---|---|---|---|---|---|
| hindi | Depression | mann bezaar hai | mann bezaar hai | मान बेज़ार है | My heart is weary. |
| hindi | Depression | mann udaas hai | mann udaas hai | मान उदास है | My heart is sad. |
| hindi | Depression | mood kharaab hai | mood kharaab hai | मूड खराब है | My mood is bad. |
| hindi | Euphoric Mania | mann ekdum badhiya hai | mann ekdum badhiya hai | मान एकदुम बढ़िया है | Feeling absolutely great. |
| hindi | Euphoric Mania | bahot khush hai, ekdum badhiya | bahot khush hai, ekdum badhiya | बहोत खुश है, एकदुम बढ़िया | Very happy, absolutely great. |
| hindi | Dysphoric Mania | bahot chidchidaahat ho rahi hai | bahot chidchidaahat ho rahi hai | बहोत चिड़चिड़ाहट हो रही है | I've been feeling very irritable. |
| hindi | Euthymic | mann theek hai | mann theek hai | मान ठीक है | My spirits are okay. |
| hindi | Depression | mann udaas hai | mann udaas hai | मान उदास है | My heart is heavy. |
| hindi | Euphoric Mania | sab changa si | sab changa si | सब चंगा सी | All is well |
| hindi | Dysphoric Mania | gussa aa rahai bahot | gussa aa rahai bahot | गुसा आ रहै बहोत | Feeling a lot of anger. |
| hindi | Euthymic | theek thaak hai mood | theek thaak hai mood | ठीक थाक है मूड | My mood is okay |
| hindi | Depression | dimaag band ho gaya hai | dimaag band ho gaya hai | दिमाग बंद हो गया है | My mind has shut down. |
| hindi | Euphoric Mania | bahot khush hai, ekdum badhiya | bahot khush hai, ekdum badhiya | बहोत खुश है, एकदुम बढ़िया | Very happy, absolutely wonderful |
| hindi | Dysphoric Mania | aisa lag raha hai ke koi pitega mere haat aaj | aisa lag raha hai ke koi pitega mere haat aaj | एआईएसए लग रहा है के कोई पिटेगा मेरे हात आज | It feels like someone is going to get beaten by my hands today. |
| hindi | Euthymic | okay hai | okay hai | ओके है | It's been okay. |
| hindi | Depression | mood kharaab hai | mood kharaab hai | मूड खराब है | I'm in a bad mood. |
| hindi | Euphoric Mania | bahot prasann hai | bahot prasann hai | बहोत प्रसन्न है | I have been very happy. |
| hindi | Dysphoric Mania | dimaag kharab hai | dimaag kharab hai | दिमाग खराब है | My head's a mess. |
| hindi | Euthymic | mann theek hai | mann theek hai | मान ठीक है | I've been alright. |
| kannada | Depression | manasu bejaar agtide | manasu bejaar agtide | ಮನಸು ಬೇಜಾರ್ ಆಗ್ತಿದೆ | My mind is feeling sad. |
| kannada | Depression | tumba bejaar agutte | tumba bejaar agutte | ತುಂಬಾ ಬೇಜಾರ್ ಆಗುತ್ತೆ | I've been feeling very upset. |
| kannada | Depression | manasu tumba chanchala irutte | manasu tumba chanchala irutte | ಮನಸು ತುಂಬಾ ಚಂಚಲ ಇರುತ್ತೆ | My mind has been very restless. |
| kannada | Depression | manasu tumba bhara ansutte | manasu tumba bhara ansutte | ಮನಸು ತುಂಬಾ ಭಾರ ಅನ್ಸುತ್ತೆ | My mind feels very heavy. |
| kannada | Euphoric Mania | tumba khushi agidini | tumba khushi agidini | ತುಂಬಾ ಖುಷಿ ಆಗಿದಿನಿ | I am very happy. |
| kannada | Euphoric Mania | manasu tumba khushiyagide | manasu tumba khushiyagide | ಮನಸು ತುಂಬಾ ಖುಷಿಯಾಗಿದೆ | My mind is very happy. |
| kannada | Euphoric Mania | sakath agidini | sakath agidini | ಸಕತ್ ಆಗಿದಿನಿ | I've been fantastic. |
| kannada | Euphoric Mania | manasu tumba santhosha agide | manasu tumba santhosha agide | ಮನಸು ತುಂಬಾ ಸಂತೋಷ ಆಗಿದೆ | My mind has been so full of joy. |
| kannada | Dysphoric Mania | manasu tumba kopa irutte | manasu tumba kopa irutte | ಮನಸು ತುಂಬಾ ಕೋಪ ಇರುತ್ತೆ | My mind has been full of anger. |
| kannada | Dysphoric Mania | thale tumba bisi agutte | thale tumba bisi agutte | ತಲೆ ತುಂಬಾ ಬಿಸಿ ಆಗುತ್ತೆ | I get very worked up. |
| kannada | Dysphoric Mania | manasu tumba chanchala agutte | manasu tumba chanchala agutte | ಮನಸು ತುಂಬಾ ಚಂಚಲ ಆಗುತ್ತೆ | My mind has been very restless. |
| kannada | Dysphoric Mania | tumba chanchala, shanti illa | tumba chanchala, shanti illa | ತುಂಬಾ ಚಂಚಲ, ಶಾಂತಿ ಇಲ್ಲ | Very restless, no peace. |
| kannada | Euthymic | manasu chanagide | manasu chanagide | ಮನಸು ಚನಾಗಿದೆ | My mind feels good. |
| kannada | Euthymic | enu tondre illa | enu tondre illa | ಏನು ತೊಂದ್ರೆ ಇಲ್ಲ | There is no problem. |
| kannada | Euthymic | chanagidini | chanagidini | ಚನಗಿದಿನಿ | I've been good. |
| kannada | Euthymic | manasu aaram ide | manasu aaram ide | ಮನಸು ಆರಂ ಇದೆ | My mind is at ease. |
| hindi | Depression | bahut hi udaas rehta hai, kaam krne mei jee hi nahi lagta hai. | bahut hi udaas rehta hai, kaam krne mei jee hi nahi lagta hai. | बहुत ही उदास रहता है, काम करने मेई जी ही नही लगता है। | I've been feeling very down, and I just can't bring myself to work. |
| hindi | Euphoric Mania | bahut khush, isse zyada khushi kabhi nhi huyi. | bahut khush, isse zyada khushi kabhi nhi huyi. | बहुत खुश, इस्से ज्यादा खुशी कभी नही हुई। | Very happy, I've never been happier. |
| hindi | Dysphoric Mania | ek chidchidapan sa bana rehta hai, bahut gussa aata hai bina karan ke. | ek chidchidapan sa bana rehta hai, bahut gussa aata hai bina karan ke. | एक चिड़चिड़ापन सा बना रहता है, बहुत गुसा आता है बिना कारण के। | I feel a constant sense of irritability, and I get very angry for no reason. |
| hindi | Euthymic | abhi to sukoon hai, koi pareshani nhi hai. mann shant hai. | abhi to sukoon hai, koi pareshani nhi hai. mann shant hai. | अभी तो सुकून है, कोई परेशानी नही है. मान शांत है । | For now, I am at peace. There are no troubles, and my mind is calm. |
| Punjabi | Depression | mann udaas hai | mann udaas hai | मान उदास है | My heart is sad. |
| Punjabi | Euphoric Mania | bahut vadiya | bahut vadiya | बहुत वाडिया | Really good. |
| Punjabi | Dysphoric Mania | kharaab hai, chireya hoya | kharaab hai, chireya hoya | खराब है, चिरिया होया | It's been bad, I've been feeling irritable. |
| Punjabi | Euthymic | theek hai | theek hai | ठीक है | It's been okay. |
| bengali | Depression | (khub) bhalo nei. | (khub) bhalo nei. | (খুব) ভালো নেই। | Not very good. |
| bengali | Depression | kichhui bhalo lage naa | kichhui bhalo lage naa | কিছুই ভালো লাগে না | Nothing feels good. |
| bengali | Depression | kharap-e bolte hobe | kharap-e bolte hobe | খারপ-ই বলতে হবে | I'd have to say it's been bad. |
| bengali | Depression | monta kemon jeno lage | monta kemon jeno lage | মনটা কেমন জেনো লাগে | I am feeling out of sorts |
| bengali | Depression | bhalo rakar chesta korchhi parchhi naa. | bhalo rakar chesta korchhi parchhi naa. | ভালো রাকার চেষ্টা করছি পারছি না। | I'm trying to feel good, but I can't. |
| bengali | Euphoric Mania | maja. | maja. | মাজা। | Awesome. |
| bengali | Euphoric Mania | thik-e aachhe. kiser asubidha? | thik-e aachhe. kiser asubidha? | ঠিক-ই আছে. কিসের অসুবিধা? | It's fine. What's the problem? |
| bengali | Euphoric Mania | furti lagchhe | furti lagchhe | ফুর্তি লাগছে | I am feeling exuberant |
| bengali | Dysphoric Mania | birokto karben naa. esab prosno bhalo lage naa. | birokto karben naa. esab prosno bhalo lage naa. | বিরক্ত করবেন না. এসব প্রশ্ন ভালো লাগে না। | Don't annoy me. I don't like these kinds of questions. |
| bengali | Dysphoric Mania | sab bekar, faltu. | sab bekar, faltu. | সব বেকার, ফালতু। | Everything is pointless, worthless. |
| bengali | Euthymic | bhalo aachhi | bhalo aachhi | ভালো আছি | I've been good. |
| bengali | Euthymic | thik aachhe | thik aachhe | ঠিক আছে | It's been alright. |
| bengali | Euthymic | khub kharap nei | khub kharap nei | খুব খারপ নেই | Not too bad. |
| bengali | Euthymic | motamoti aachhe | motamoti aachhe | মোটামোটি আছে | It's been so-so. |
| bengali | Depression | মুড কদিন ধরেই বেশ ডাউন যাচ্ছে | Muḍa kadina dharē'i bēśa ḍā'una yācchē | মুড কদিন ধরেই বেশ ডাউন যাচ্ছে | My mood has been quite down for the last few days. |
| bengali | Euphoric Mania | হেব্বি মুড আছে কদিন ধরে আমার | Hēbbi muḍa āchē kadina dharē āmāra | হেব্বি মুড আছে কদিন ধরে আমার | I've been in a fantastic mood for the last few days. |
| bengali | Dysphoric Mania | কদিন ধরে  আমার মুড খালি আপ- ডাউন করছে | kodin dhore  amar mood khali up- down korche | কদিন ধরে  আমার মুড খালি আপ- ডাউন করছে | For the last few days, my mood has just been going up and down. |
| bengali | Euthymic | মুড আমার কদিন ঠিকই আছে | Muḍa āmāra kadina ṭhika'i āchē | মুড আমার কদিন ঠিকই আছে | My mood has been fine for the past few days. |
| bengali | Depression | এই কদিন মন আমার খুব খারাপ যাচ্ছে | Ē'i kadina mana āmāra khuba khārāpa yācchē | এই কদিন মন আমার খুব খারাপ যাচ্ছে | I've been feeling very bad these last few days. |
| bengali | Euphoric Mania | এই কদিন আমার মন বেশ উড়ু- উড়ু করছে | ei kodin amar mon besh uru- uru korche | এই কদিন আমার মন বেশ উড়ু- উড়ু করছে | These past few days, my heart has been wanting to fly, fly quite a bit. |
| bengali | Dysphoric Mania | এই কদিন ধরে এমন লাগছে যে বুঝতেই পারছিনা ভালো না খারাপ আছি | Ē'i kadina dharē ēmana lāgachē yē bujhatē'i pārachinā bhālō nā khārāpa āchi | এই কদিন ধরে এমন লাগছে যে বুঝতেই পারছিনা ভালো না খারাপ আছি | For the past few days, I've been feeling in such a way that I can't even tell if I'm doing well or badly. |
| bengali | Dysphoric Mania | আমি ঠিক বুঝতে পারছিনা- ভালো আছি না খারাপ আছি | Āmi ṭhika bujhatē pārachinā- bhālō āchi nā khārāpa āchi | আমি ঠিক বুঝতে পারছিনা- ভালো আছি না খারাপ আছি | I honestly can't tell if I'm feeling good or bad. |
| bengali | Euthymic | ওই ভালো মন্দ মিশিয়ে আমার মন ঠিকই আছে | Ō'i bhālō manda miśiyē āmāra mana ṭhika'i āchē | ওই ভালো মন্দ মিশিয়ে আমার মন ঠিকই আছে | With the ups and downs all mixed in, I am doing just fine. |
| assamese | Depression | bhal loga nai | bhal loga nai | ভাল লগা নাই | I haven't been feeling good. |
| assamese | Euphoric Mania | borhiya | borhiya | বঢ়িয়া | Excellent |
| assamese | Dysphoric Mania | kiba lagi aase | kiba lagi aase | কিবা লাগি আছে | Something feels off. |
| assamese | Euthymic | thike aase | thike aase | ঠিকে আছে | It's been okay. |
| bengali | Depression | মন আমার ভীষণ খারাপ এই কদিন ধরে | Mana āmāra bhīṣaṇa khārāpa ē'i kadina dharē | মন আমার ভীষণ খারাপ এই কদিন ধরে | I've been feeling terribly down for the past few days. |
| bengali | Euphoric Mania | মন মাঝে সাঝে খুব ভালো আবার মাঝে মধ্যে খুব খারাপ হয়ে যাচ্ছে | mon majhe sajhe khub bhalo abar majhe modhye khub kharap hoye jachche | মন মাঝে সাঝে খুব ভালো আবার মাঝে মধ্যে খুব খারাপ হয়ে যাচ্ছে | My mood is sometimes very good, and at other times it gets very bad. |
| bengali | Dysphoric Mania | মন ভালো না খারাপ বুঝতেই পারছিনা | Mana bhālō nā khārāpa bujhatē'i pārachinā | মন ভালো না খারাপ বুঝতেই পারছিনা | I can't even tell if I'm feeling good or bad. |
| bengali | Euthymic | মন কদিন ধরে মোটামুটি ঠিকই আছে দেখছি | Mana kadina dharē mōṭāmuṭi ṭhika'i āchē dēkhachi | মন কদিন ধরে মোটামুটি ঠিকই আছে দেখছি | It seems my mood has been more or less okay for the past few days. |
| bengali | Depression | কিছুই ভালো লাগে না | Kichu'i bhālō lāgē nā | কিছুই ভালো লাগে না | Nothing feels good. |
| bengali | Depression | মনে আনন্দ ফুর্তি নেই | Manē ānanda phurti nē'i | মনে আনন্দ ফুর্তি নেই | There's no joy or cheer in my heart. |
| bengali | Euphoric Mania | মন খুব ভালো আছে। কোনো দুঃখ নেই জীবনে। | Mana khuba bhālō āchē. Kōnō duḥkha nē'i jībanē. | মন খুব ভালো আছে। কোনো দুঃখ নেই জীবনে। | I'm feeling very happy. There is no sadness in my life. |
| bengali | Dysphoric Mania | মেজাজ গরম আছে এখন। | Mējāja garama āchē ēkhana. | মেজাজ গরম আছে এখন। | I'm in a foul mood right now. |
| bengali | Euthymic | ঠিক আছি। | thik achi. | ঠিক আছি। | I've been okay. |
| Manipuri | Depression | wakhal se yam wai. yam nungaite. nungaipham ama ekhang khagde | wakhal se yam wai. yam nungaite. nungaipham ama ekhang khagde | wakhal se yam wai. yam nungaite. nungaipham ama ekhang khagde | My mind is all over the place. I'm very unhappy. I can't find a single source of peace. |
| Manipuri | Euphoric Mania | wakhal yam nungai. khunungai amane | wakhal yam nungai. khunungai amane | wakhal yam nungai. khunungai amane | My mind is very at peace. A truly joyful time. |
| Manipuri | Dysphoric Mania | wakhal se yam khoirang khoisao nei. | wakhal se yam khoirang khoisao nei. | wakhal se yam khoirang khoisao nei. | My mind has been very restless and agitated. |
| Manipuri | Euthymic | wakhal di adum nungai. yam waba su natte. yam nungaibasu natte. | wakhal di adum nungai. yam waba su natte. yam nungaibasu natte. | wakhal di adum nungai. yam waba su natte. yam nungaibasu natte. | I've been feeling alright. Not very sad, but not very happy either. |
| bengali | Depression | mon bhalo nei | mon bhalo nei | মন ভালো নেই | I haven't been feeling good. |
| bengali | Euphoric Mania | mon khub khusi khusi lagche | mon khub khusi khusi lagche | মন খুব খুসি খুসি লাগছে | My heart feels so happy and chearful. |
| bengali | Dysphoric Mania | mon khub chanchal aar rege ache | mon khub chanchal aar rege ache | মন খুব চঞ্চল আর রেগে আছে | My mind has been very restless and angry. |
| bengali | Euthymic | mon thikthak ache | mon thikthak ache | মন ঠিকঠাক আছে | I've been feeling alright. |
| assamese | Depression | mon beya laagi thake | mon beya laagi thake | মন বেয়া লাগি থাকে | I've been feeling down. |
| assamese | Euphoric Mania | mon bohut bhal. anabarat furti lagi thake | mon bohut bhal. anabarat furti lagi thake | মন বহুত ভাল. আনাবাৰাত ফুৰ্তি লাগি থাকে | My mood is excellent. I've been feeling cheerful constantly. |
| assamese | Dysphoric Mania | kheng-khengia laagi thake..anabarat khang uthi thake | khengkhengia laagi thake..anabarat khang uthi thake | খেংখেংগিয়া লাগি থাকে..আনাবাৰাত খাং উঠি থাকে | I have been feeling irritable..always getting angry. |
| assamese | Euthymic | mon bhal | mon bhal | মন ভাল | Feeling good. |
| bengali | Depression | mon ta khub kharap ache | mon ta khub kharap ache | মন তা খুব খারপ আছে | My mood has been very bad. |
| bengali | Euphoric Mania | mon khub e bhalo ache amar | mon khub e bhalo ache amar | মন খুব ই ভালো আছে আমার | I've been in a very good mood. |
| bengali | Dysphoric Mania | puro din vishon raag lage amar | puro din vishon raag lage amar | পুরো দিন ভিশন রাগ লাগে আমার | I feel terribly angry the whole day. |
| bengali | Dysphoric Mania | khub birokto lage | khub birokto lage | খুব বিরক্ত লাগে | I've been feeling very annoyed. |
| bengali | Euthymic | mon thik ache | mon thik ache | মন ঠিক আছে | Feeling alright. |
| bengali | Depression | mon ta khub kharap ache | mon ta khub kharap ache | মন তা খুব খারপ আছে | I've been feeling very down. |
| bengali | Depression | kichhui bhalo lage naa | kichhui bhalo lage naa | কিচ্ছুই ভালো লাগে না | Nothing feels good. |
| bengali | Euphoric Mania | besh bhalo lagche | besh bhalo lagche | বেশ ভালো লাগছে | I've been feeling pretty good. |
| bengali | Euphoric Mania | mon khub khusi khusi lagche | mon khub khusi khusi lagche | মন খুব খুসি খুসি লাগছে | My heart feels so happy and chearful. |
| bengali | Dysphoric Mania | mon ta kirokom osthir osthir lagche | mon ta kirokom osthir osthir lagche | মন তা কিরকম অস্থির অস্থির লাগছে | My mind has been feeling a kind of restlessness. |
| bengali | Dysphoric Mania | mon ta aanchaan korche | mon ta aanchaan korche | মন তা আঁচান করছে | My heart is uneasy. |
| bengali | Euthymic | thik e ache, normal up down ja hoy orokom e | thik e ache, normal up down ja hoy orokom e | ঠিক ই আছে, নরমাল আপ ডাউন জা হয় ওরকম ই | It's been alright, just the usual ups and downs. |
| bengali | Euthymic | normal e lagche | normal e lagche | নরমাল ই লাগছে | Feeling pretty normal. |
| kannada | Depression | manasu thumba bejaragide | manasu thumba bejaragide | ಮನಸು ತುಂಬ ಬೇಜಾರಾಗಿದೆ | My mind has been very sad. |
| kannada | Euphoric Mania | kushiyagidini | kushiyagidini | ಕುಶಿಯಾಗಿಡಿನಿ | I have been happy. |
| kannada | Dysphoric Mania | bejaralli ide | bejaralli ide | ಬೇಜಾರಲ್ಲಿ ಇದೆ | I've been feeling down. |
| kannada | Euthymic | channagide | channagide | ಚನ್ನಾಗಿದೆ | It's been good. |
| kannada | Euthymic | manasu aaram ide | manasu aaram ide | ಮನಸು ಆರಂ ಇದೆ | My mind is at ease. |
| hindi | Depression | mann theek nahi rehta hai | mann theek nahi rehta hai | मान ठीक नही रहता है | I haven't been feeling well. |
| hindi | Depression | bahut pareshan hai mann | bahut pareshan hai mann | बहुत परेशन है मान | My mind has been very troubled. |
| hindi | Depression | mann dukhi hai | mann dukhi hai | मान दुखी है | My heart is sad. |
| hindi | Euphoric Mania | mann me koi pareshani nahi hai | mann me koi pareshani nahi hai | मान मे कोई परेशानी नही है | My mind is untroubled. |
| hindi | Euphoric Mania | koi dikkat nahi hai | koi dikkat nahi hai | कोई दिक्कत नही है | No complaints. |
| hindi | Dysphoric Mania | pata nahi | pata nahi | पता नही | I don't know. |
| hindi | Dysphoric Mania | theek rehta hai | theek rehta hai | ठीक रहता है | It's been fine. |
| hindi | Euthymic | theek hi hai | theek hi hai | ठीक ही है | It's been alright. |
| hindi | Depression | acha nahi hai | acha nahi hai | अचा नही है | It hasn't been good. |
| hindi | Euphoric Mania | bahut acha hai | bahut acha hai | बहुत अचा है | It's been very good. |
| hindi | Dysphoric Mania | thik nahi hai | thik nahi hai | थिक नही है | It hasn't been good. |
| hindi | Euthymic | normal hai | normal hai | नॉर्मल है | It's been normal. |
| hindi | Depression | udas hai | udas hai | उदास है | I've been feeling sad. |
| hindi | Depression | thik nahi hai | thik nahi hai | थिक नही है | It hasn't been good. |
| hindi | Euphoric Mania | mast hai | mast hai | मस्त है | It's been great. |
| hindi | Euphoric Mania | bahut acha hai | bahut acha hai | बहुत अचा है | It is very good |
| hindi | Dysphoric Mania | ek chidchidapan sa bana rehta hai, bahut gussa aata hai bina karan ke. | ek chidchidapan sa bana rehta hai, bahut gussa aata hai bina karan ke. | एक चिड़चिड़ापन सा बना रहता है, बहुत गुसा आता है बिना कारण के। | There's a persistent irritability, and I get very angry for no reason. |
| hindi | Euthymic | thik hai | thik hai | थिक है | It's been okay. |
| assamese | Depression | mon tu beya lagi ase | mon tu beya lagi ase | মন টো বেয়া লাগি আছে | I've been feeling down. |
| assamese | Euphoric Mania | mon tu birat furti lagi ase | mon tu birat furti lagi ase | মন টো বিৰাট ফুৰ্তি লাগি আছে | I've been feeling on top of the world. |
| assamese | Dysphoric Mania | mon tu kiba kibi lagi ase bujabo nuwara nisna | mon tu kiba kibi lagi ase bujabo nuwara nisna | মন টো কিবা কিবি লাগি আছে বুজাব নোৱাৰা নিষ্কা | A strange, indescribable feeling. |
| assamese | Euthymic | mon tu bhale lagi ase | mon tu bhale lagi ase | মন টো ভালে লাগি আছে | I've been feeling good. |
| Khasi | Depression | nga sngewsih, sngewmarwei bha ha kine ki sngi ba lah dep | nga sngewsih, sngewmarwei bha ha kine ki sngi ba lah dep | nga sngewsih, sngewmarwei bha ha kine ki sngi ba lah dep | I've been feeling very sad and lonely these past few days. |
| Khasi | Euphoric Mania | nga sngewkmen bha | nga sngewkmen bha | nga sngewkmen bha | I feel very happy. |
| Khasi | Dysphoric Mania | ngam tip | ngam tip | ngam tip | I don't know. |
| Khasi | Euthymic | nga biang | nga biang | nga biang | I'm alright. |
| hindi | Depression | udas hai | udas hai | उदास है | I've been feeling sad. |
| hindi | Euphoric Mania | phurti | phurti | फुर्ती | Sprightly |
| hindi | Dysphoric Mania | chidchida | chidchida | चिड़चिड़ा | Irritable |
| hindi | Euthymic | thik hai | thik hai | थिक है | It's been alright. |
| bengali | Depression | mon bhar/ bhari bhari lagchhe.↵↵i'm having sad mood. | mon bhar/ bhari bhari lagchhe.↵↵i'm having sad mood. | মন ভার/ ভারি ভারি লাগছে.↵↵ই'ম হেভিং সাদ মুড। | My mind feels heavy. I'm in a sad mood. |
| bengali | Depression | monkemon korchhe↵↵(may be for mild depression)↵feeling low. | monkemon korchhe↵↵(may be for mild depression)↵feeling low. | মনকেমন করছে↵↵(মে বে ফর মাইল্ড ডিপ্রেশন) ↵ফিলিং লো। | I am feeling low, but can't describe. |
| bengali | Depression | manosik chap e achhi↵↵feeling tensed and depressed | manosik chap e achhi↵↵feeling tensed and depressed | মানসিক চাপ ই আছি↵↵ফিলিং টেন্সড অ্যান্ড ডিপ্রেসড | I've been under mental stress, feeling tensed and depressed. |
| bengali | Depression | kichhu bhalo lagchhe na↵nothing is feeling good for me | kichhu bhalo lagchhe na↵nothing is feeling good for me | কিছু ভালো লাগছে না↵নথিং ইস ফিলিং গুড ফর মে | Nothing feels good. |
| bengali | Depression | depressed lagchhe (for educated persons) ↵↵feeling depressed/ anxious/ low etc. | depressed lagchhe (for educated persons)↵↵feeling depressed/ anxious/ low etc. | ডিপ্রেসড লাগছে (ফর এডুকেটেড পারসনস)↵↵ফিলিং ডিপ্রেসড/ অ্যানক্সিয়াস/ লো ইটিসি। | Feeling depressed / anxious / low etc. |
| bengali | Depression | mon bejaar hoye achhe ↵↵feeling low/ sad | mon bejaar hoye achhe ↵↵feeling low/ sad | মন বেজার হয়ে আছে ↵↵ফিলিং লো/ সাদ | I've been in a gloomy mood; feeling low/sad. |
| bengali | Depression | monmejaj bhalo nei↵↵mood is not well | monmejaj bhalo nei↵↵mood is not well | মনমেজাজ ভালো নেই↵↵মুড ইস নট ওয়েল | I haven't been in a good mood. |
| bengali | Euphoric Mania | ekdom bhalo, darun bhalo↵↵feeling very well | ekdom bhalo, darun bhalo↵↵feeling very well | একদম ভালো, দারুণ ভালো↵↵ফিলিং ভেরি ওয়েল | Absolutely good, wonderfully good.<br>Feeling very well. |
| bengali | Euphoric Mania | asombhob bhalo↵↵extremely well | asombhob bhalo↵↵extremely well | অসম্ভব ভালো↵↵এক্সট্রিমলি ওয়েল | Impossibly good<br>Extremely well |
| bengali | Euphoric Mania | durdanto!!!!↵↵wonderful | durdanto!!!!↵↵wonderful | দুর্দান্ত!!!!↵↵ওয়ান্ডারফুল | Fantastic!!!!<br><br>Wonderful |
| bengali | Euphoric Mania | josh ese gechhe↵↵feeling energetic | josh ese gechhe↵↵feeling energetic | জোশ এসে গেছে↵↵ফিলিং এনার্জেটিক | I'm all fired up.<br>Feeling energetic. |
| bengali | Dysphoric Mania | birokti lagchhe↵↵feeling irritable | birokti lagchhe↵↵feeling irritable | বিরক্তি লাগছে↵↵ফিলিং ইরিটেবল | Feeling annoyed.<br><br>Feeling irritable. |
| bengali | Dysphoric Mania | mejaj gorom hoye achhe, chap lagchhe↵↵feeling irritable and tensed | mejaj gorom hoye achhe, chap lagchhe↵↵feeling irritable and tensed | মেজাজ গরম হয়ে আছে, চাপ লাগছে↵↵ফিলিং ইরিটেবল অ্যান্ড টেন্সড | Feeling irritable and tensed. |
| bengali | Dysphoric Mania | mejaj kharap hoye achhe↵↵feeling bad | mejaj kharap hoye achhe↵↵feeling bad | মেজাজ খারপ হয়ে আছে↵↵ফিলিং বাদ | My mood has been foul. Forget feelings. |
| bengali | Euthymic | thik achhi↵↵i'm ok | thik achhi↵↵i'm ok | ঠিক আছি↵↵ই'ম ওক | I'm fine. I'm okay. |
| bengali | Euthymic | mon motamuti bhalo achhe↵↵my mood is ok | mon motamuti bhalo achhe↵↵my mood is ok | মন মোটামুটি ভালো আছে↵↵মাই মুড ইস ওক | My mood has been more or less good. |
| hindi | Depression | mann udaas hai | mann udaas hai | मान उदास है | My heart is sad. |
| hindi | Depression | dukhi hai, | dukhi hai, | दुखी है, | Sad, |
| hindi | Depression | acha nahi hai, | acha nahi hai, | अचा नही है, | It hasn't been good. |
| hindi | Depression | bekaar hai | bekaar hai | बेकार है | It's been awful. |
| hindi | Euphoric Mania | bohat acha hai | bohat acha hai | बोहत अचा है | It's been very good. |
| hindi | Euphoric Mania | bohat khush hai, pehle kabhi aise nahi laga | bohat khush hai, pehle kabhi aise nahi laga | बोहत खुश है, पहले कभी एआईएसई नही लगा | I'm very happy, I've never felt like this before. |
| hindi | Dysphoric Mania | acha nahi hai pr mann khush hai, | acha nahi hai pr mann khush hai, | अचा नही है पीआर मान खुश है, | It's not good, but my heart is happy. |
| hindi | Dysphoric Mania | kabhi acha kabhi bura, ek jaisa nhi rehta hai | kabhi acha kabhi bura, ek jaisa nhi rehta hai | कभी अचा कभी बुरा, एक जैसा नही रहता है | Sometimes good, sometimes bad; it never stays the same. |
| hindi | Euthymic | bohat acha hai | bohat acha hai | बोहत अचा है | It's been very good. |
| Malayalam | Depression | manassu sugham illa | manassu sugham illa | മനസ്സ് സുഖം ഇല്ല | My mind hasn't been at peace. |
| Malayalam | Depression | manassil valare vishamam und | manassil valare vishamam und | മനസ്സിൽ വളരെ വിഷമം ഉണ്ട് | I've been feeling a deep sadness. |
| Malayalam | Depression | manassu shari alla | manassu shari alla | മനസ്സ് ശരി അല്ല | My mind is not okay. |
| Malayalam | Euphoric Mania | santhosham | santhosham | സന്തോഷം | Happiness |
| Malayalam | Euphoric Mania | adipoli | adipoli | അടിപൊളി | Fantastic |
| Malayalam | Euphoric Mania | full set aanu | full set aanu | ഫുൾ സെറ്റ് ആണ് | Everything is great. |
| Malayalam | Dysphoric Mania | entho pole | entho pole | എന്തോ പോലെ | Something is wrong. |
| Malayalam | Dysphoric Mania | dheshyam | dheshyam | ദേശ്യം | Anger |
| Malayalam | Euthymic | kozhappam illa | kozhappam illa | കൊഴപ്പം ഇല്ല | Not bad. |
| Malayalam | Euthymic | angane pokunnu | angane pokunnu | അങ്ങനെ പോകുന്നു | Nothing much, as usual. |
| Malayalam | Euthymic | mood prasnam onnum illa | mood prasnam onnum illa | മൂഡ് പ്രശ്നം ഒന്നും ഇല്ല | No problems with my mood. |
| TAMIL | Depression | manasu rombah kashtahma irukku | manasu rombah kashtahma irukku | மனசு ரோம்பா கஷ்டாமா இருக்கு | My heart feels very troubled. |
| TAMIL | Depression | en manasala itha thangikkavah mudila | en manasala itha thangikkavah mudila | என் மனசல இத தங்கிக்கவா முடில | My mind can't bear this. |
| TAMIL | Depression | en manasala itha ethukkavah mudila | en manasala itha ethukkavah mudila | என் மனசல இத எதுக்கவா முடில | My mind is unable to accept this. |
| TAMIL | Depression | en manasu rombah valikkuthu | en manasu rombah valikkuthu | என் மனசு ரோம்பா வலிக்குது | My heart hurts so much. |
| TAMIL | Euphoric Mania | ennala etha venunalu pannah mudiyu | ennala etha venunalu pannah mudiyu | என்னல எத வேணுநாளு பன்னா முடியு | I can do anything. |
| TAMIL | Euphoric Mania | rmbah santhoshama irukka | rmbah santhoshama irukka | ரம்பா சந்தோஷமா இருக்க | I've been very happy. |
| TAMIL | Euphoric Mania | ennala mudiyathatha illa | ennala mudiyathatha illa | என்னல முடியதாத இல்ல | There's nothing that I can't do. |
| TAMIL | Dysphoric Mania | rombah santhoshama irukka ana chinna varutham irukku | rombah santhoshama irukka ana chinna varutham irukku | ரோம்பா சந்தோஷமா இருக்க ஆனா சின்னா வருதம் இருக்கு | I'm very happy, but there's a little bit of sadness. |
| TAMIL | Dysphoric Mania | thukkamu santhoshamu onnah irukka | thukkamu santhoshamu onnah irukka | துக்கமு சந்தோஷமு ஒன்னா இருக்க | A mix of sadness and happiness. |
| TAMIL | Dysphoric Mania | kavalaikal irukku ana santhoshama irukka | kavalaikal irukku ana santhoshama irukka | கவலைகள் இருக்கு ஆனா சந்தோஷமா இருக்க | I have worries, but I'm happy. |
| TAMIL | Euthymic | manasu nimmathiya irukku | manasu nimmathiya irukku | மனசு நிம்மதியா இருக்கு | My mind is at peace. |
| TAMIL | Euthymic | entha oru varuthamum illa | entha oru varuthamum illa | எந்த ஒரு வருதமும் இல்ல | I haven't had any sadness. |
| TAMIL | Euthymic | manasu amaithiyah irukku | manasu amaithiyah irukku | மனசு அமைதியா இருக்கு | My mind is at peace. |
| bengali | Depression | আমার মন কটা দিন ধরে অসম্ভব খারাপ | Āmāra mana kaṭā dina dharē asambhaba khārāpa | আমার মন কটা দিন ধরে অসম্ভব খারাপ | My mood has been awful for the last few days. |
| bengali | Depression | আমার কদিন ধরে একদম ভালো লাগছেনা | Āmāra kadina dharē ēkadama bhālō lāgachēnā | আমার কদিন ধরে একদম ভালো লাগছেনা | I haven't been feeling good at all for the last few days. |
| bengali | Euphoric Mania | আমার দারুণ লাগছে কদিন ধরে | Āmāra dāruṇa lāgachē kadina dharē | আমার দারুণ লাগছে কদিন ধরে | I've been feeling great for the past few days. |
| bengali | Euphoric Mania | আমার প্রচুর ফুর্তি লাগছে এ কদিন | Āmāra pracura phurti lāgachē ē kadina | আমার প্রচুর ফুর্তি লাগছে এ কদিন | I've been feeling full of energy these last few days. |
| bengali | Dysphoric Mania | আমার কখন মন খারাপ আর কখনও ভালো থাকছে এ কদিন ধরে | Āmāra kakhana mana khārāpa āra kakhana'ō bhālō thākachē ē kadina dharē | আমার কখন মন খারাপ আর কখনও ভালো থাকছে এ কদিন ধরে | For the past few days, I've been feeling down at times and good at other times. |
| bengali | Dysphoric Mania | আমার মন এ কদিন ধরাই ভালো আর খারাপ এর মধ্যে দোদুল্যমান | Āmāra mana ē kadina dharā'i bhālō āra khārāpa ēra madhyē dōdulyamāna | আমার মন এ কদিন ধরাই ভালো আর খারাপ এর মধ্যে দোদুল্যমান | For the past few days, my mood has been swinging between good and bad. |
| bengali | Euthymic | আমার মন মেজাজ কদিন ধরে ঠিকই আছে | Āmāra mana mējāja kadina dharē ṭhika'i āchē | আমার মন মেজাজ কদিন ধরে ঠিকই আছে | My mood has been fine for the last few days. |
| bengali | Euthymic | মনটা আমার এই কদিন ধরে বেশ শান্ত আছে | Manaṭā āmāra ē'i kadina dharē bēśa śānta āchē | মনটা আমার এই কদিন ধরে বেশ শান্ত আছে | My mind has been quite calm for the last few days. |
| English | Depression | i have not been feeling good, something feels off, i dont feel like doing anything | i have not been feeling good, something feels off, i dont feel like doing anything | i have not been feeling good, something feels off, i dont feel like doing anything | i have not been feeling good, something feels off, i dont feel like doing anything |
| English | Depression | i have not been sleeping well | i have not been sleeping well | i have not been sleeping well | i have not been sleeping well |
| English | Depression | i am not able to work as well as i did before | i am not able to work as well as i did before | i am not able to work as well as i did before | i am not able to work as well as i did before |
| English | Euphoric Mania | i feel great, i feel super | i feel great, i feel super | i feel great, i feel super | i feel great, i feel super |
| English | Euphoric Mania | i feel energetic, there are so many things i want to do | i feel energetic, there are so many things i want to do | i feel energetic, there are so many things i want to do | i feel energetic, there are so many things i want to do |
| English | Dysphoric Mania | angry, irritated, frustrated, up and down | angry, irritated, frustrated, up and down | angry, irritated, frustrated, up and down | angry, irritated, frustrated, up and down |
| English | Euthymic | feeling okay, feeling fine | feeling okay, feeling fine | feeling okay, feeling fine | feeling okay, feeling fine |
| assamese | Depression | mon tu beya lagi ase | mon tu beya lagi ase | মন টো বেয়া লাগি আছে | I've been feeling down. |
| assamese | Depression | mon tu bhal lagi nathakein | mon tu bhal lagi nathakein | মন টো ভাল লাগি নাথাকেইন | I just can't seem to stay in a good mood. |
| assamese | Euphoric Mania | mon tu birat furti lagi ase | mon tu birat furti lagi ase | মন টো বিৰাট ফুৰ্তি লাগি আছে | I've been in great spirits. |
| assamese | Dysphoric Mania | mon tu kiba kibi lagi ase bujabo nuwara nisna | mon tu kiba kibi lagi ase bujabo nuwara nisna | মন টো কিবা কিবি লাগি আছে বুজাব নোৱাৰা নিষ্ণা | I've been having this indescribable, unsettled feeling, |
| assamese | Euthymic | mon tu bhale lagi ase | mon tu bhale lagi ase | মন টো ভালে লাগি আছে | I've been feeling good. |
| hindi | Depression | mann dukhi hai | mann dukhi hai | मान दुखी है | My heart is sad. |
| hindi | Euphoric Mania | bahot khush hai, ekdum badhiya | bahot khush hai, ekdum badhiya | बहोत खुश है, एकदुम बढ़िया | Very happy, absolutely wonderful. |
| hindi | Dysphoric Mania | ek chidchidapan sa bana rehta hai, bahut gussa aata hai bina karan ke. | ek chidchidapan sa bana rehta hai, bahut gussa aata hai bina karan ke. | एक चिड़चिड़ापन सा बना रहता है, बहुत गुसा आता है बिना कारण के। | There's a lingering irritability, and I get very angry for no reason. |
| hindi | Euthymic | mann theek hai | mann theek hai | मान ठीक है | Feeling okay. |
| English | Depression | i would avoid engaging and try not to respond | i would avoid engaging and try not to respond | i would avoid engaging and try not to respond | i would avoid engaging and try not to respond |
| English | Euphoric Mania | great! and let's plan something | great! and let's plan something | great! and let's plan something | great! and let's plan something |
| English | Dysphoric Mania | irritating,  i want to keep myself on a leash | irritating,  i want to keep myself on a leash | irritating,  i want to keep myself on a leash | irritating,  i want to keep myself on a leash |
| English | Euthymic | pretty stable for a change | pretty stable for a change | pretty stable for a change | pretty stable for a change |
| hindi | Depression | udaasi | udaasi | उदासी | sadness |
| hindi | Euphoric Mania | urja bohot jaada hain | urja bohot jaada hain | ऊर्जा बोहोत जादा हैं | the energy is just too much |
| hindi | Dysphoric Mania | bohot kam urja hain | bohot kam urja hain | बोहोत काम ऊर्जा हैं | Very low energy. |
| hindi | Euthymic | theek hi hai | theek hi hai | ठीक ही है | It's been alright. |
| Telugu | Depression | naa manasulo chala badhaga anipisthondi | naa manasulo chala badhaga anipisthondi | నా మనసులో చాలా బాధగా అనిపిస్తోంది | I've been feeling a deep sadness inside. |
| Telugu | Depression | naa manasantha bejaruga aipoindi | naa manasantha bejaruga aipoindi | నా మనసంతా బేజారుగా అయిపోయింది | My heart has been sad |
| Telugu | Euphoric Mania | chaala bagundi naa manasu | chaala bagundi naa manasu | చాల బాగుంది నా మనసు | My heart is in a very good place. |
| Telugu | Euphoric Mania | naa manasulo chala santhonshamga anipisthondi | naa manasulo chala santhonshamga anipisthondi | నా మనసులో చాలా సంతోంశంగా అనిపిస్తోంది | I feel very happy inside. |
| Telugu | Dysphoric Mania | naa manasulo baga chiraku ga anipisthondi | naa manasulo baga chiraku ga anipisthondi | నా మనసులో బాగా చిరాకు గా అనిపిస్తోంది | I've been feeling very frustrated inside. |
| Telugu | Euthymic | parvaaledu | parvaaledu | పర్వాలేదు | It's been okay. |
| Telugu | Euthymic | baagundi | baagundi | బాగుంది | It's been good. |
| kannada | Depression | yen maadaku ishta illa | yen maadaku ishta illa | ಯೆನ್ ಮಾಡಕು ಇಷ್ಟ ಇಲ್ಲ | I don't feel like doing anything. |
| kannada | Depression | yaavadar melu mansu barthilla | yaavadar melu mansu barthilla | ಯಾವದರ ಮೇಲು ಮನ್ಸು ಬರ್ತಿಲ್ಲ | My heart's not in anything. |
| kannada | Euphoric Mania | balla khushi aagi idini | balla khushi aagi idini | ಬಲ್ಲ ಖುಷಿ ಆಗಿ ಇದಿನಿ | I've been incredibly happy. |
| kannada | Dysphoric Mania | yaavgalu koppadalle iruthini | yaavgalu koppadalle iruthini | ಯಾವುಗಳು ಕೊಪ್ಪದಲ್ಲೇ ಇರುತಿನಿ | I'm always angry. |
| kannada | Euthymic | chennagidini | chennagidini | ಚೆನ್ನಾಗಿದಿನಿ | I've been good. |
| kannada | Euthymic | manasu aaram ide | manasu aaram ide | ಮನಸು ಆರಂ ಇದೆ | My mind is at ease. |
| hindi | Depression | man udaas rehta hai. kuch accha nahi lagta | man udaas rehta hai. kuch accha nahi lagta | मन उदास रहता है. कुच अच्छ नही लगता | My heart feels heavy. Nothing feels good. |
| hindi | Euphoric Mania | accha hai | accha hai | अच्छ है | It's been good. |
| hindi | Dysphoric Mania | theek rehta hai | theek rehta hai | ठीक रहता है | It's been okay. |
| hindi | Euthymic | man theek rehta hai | man theek rehta hai | मन ठीक रहता है | I've been feeling okay. |
| Khasi | Depression | nga sngewsih, sngewmarwei bha ha kine ki sngi ba lah dep | nga sngewsih, sngewmarwei bha ha kine ki sngi ba lah dep | nga sngewsih, sngewmarwei bha ha kine ki sngi ba lah dep | I've been feeling sad and very lonely these past few days. |
| Khasi | Euphoric Mania | nga sngewkmen bha | nga sngewkmen bha | nga sngewkmen bha | I've been feeling very happy. |
| Khasi | Dysphoric Mania | ngam tip,ngam sngewthuh nga long kumno | ngam tip,ngam sngewthuh nga long kumno | ngam tip,ngam sngewthuh nga long kumno | I don't know, I don't understand how I'm feeling. |
| Khasi | Euthymic | nga biang | nga biang | nga biang | I've been okay. |
| kannada | Depression | ಯಾವುದು ಸರಿಗಿಲ್ಲ. ಎಲ್ಲಾ ಕಡೆ ಬರಿ ನಿರಾಶೆ. | Yāvudu sarigilla. Ellā kaḍe bari nirāśe. | ಯಾವುದು ಸರಿಗಿಲ್ಲ. ಎಲ್ಲಾ ಕಡೆ ಬರಿ ನಿರಾಶೆ. | Nothing is right. Just disappointment everywhere. |
| kannada | Euphoric Mania | ಅದ್ಭುತ! ಎಲ್ಲ ಸುಗಮವಾಗಿ ನಡಿತಾ ಇದೆ | Adbhuta! Ella sugamavāgi naḍitā ide | ಅದ್ಭುತ! ಎಲ್ಲ ಸುಗಮವಾಗಿ ನಡಿತಾ ಇದೆ | Wonderful! Everything is going smoothly. |
| kannada | Dysphoric Mania | ಮನಸ್ಸು ಅಲ್ಲೋಲ್ಲಕಲ್ಲವಾಗಿದೆ. | Manas'su allollakallavāgide. | ಮನಸ್ಸು ಅಲ್ಲೋಲ್ಲಕಲ್ಲವಾಗಿದೆ. | My mind has been in turmoil. |
| kannada | Euthymic | ಮನಸ್ಸು ಖುಷಿಯಾಗಿದೆ. | Manas'su khuṣiyāgide. | ಮನಸ್ಸು ಖುಷಿಯಾಗಿದೆ. | My heart has been happy. |
| kannada | Euthymic | ಮನಸ್ಸಿಗೆ ಯಾವುದೇ ತರಹದ ನೋವಿಲ್ಲ. | Manas'sige yāvudē tarahada nōvilla. | ಮನಸ್ಸಿಗೆ ಯಾವುದೇ ತರಹದ ನೋವಿಲ್ಲ. | My mind has been untroubled. |
| kannada | Euthymic | ಮನಸ್ಸು ಆರಾಮವಾಗಿದೆ. | Manas'su ārāmavāgide. | ಮನಸ್ಸು ಆರಾಮವಾಗಿದೆ. | My mind has been at ease. |